\documentclass[11pt]{article}

\usepackage[preprint]{acl}

\usepackage{times}
\usepackage{latexsym}

\usepackage[T1]{fontenc}

\usepackage[utf8]{inputenc}

\usepackage{microtype}

\usepackage{inconsolata}

\usepackage{graphicx}
\usepackage{svg}

\usepackage[utf8]{inputenc} %
\usepackage[T1]{fontenc}    %
\usepackage{hyperref}       %
\usepackage{url}            %
\usepackage{booktabs}       %
\usepackage{amsfonts}       %
\usepackage{nicefrac}       %
\usepackage{microtype}      %
\usepackage{xcolor}         %

\usepackage{ booktabs, tabularx, siunitx}

\usepackage{microtype}
\usepackage{hyperref}
\usepackage{url}
\usepackage{booktabs}
\usepackage[most]{tcolorbox}
\usepackage{listings}
\usepackage{xspace}
\usepackage{cleveref}
\usepackage{fontawesome5}
\usepackage{wrapfig}
\usepackage{graphicx}
\usepackage{amsmath,amssymb,graphicx}
\usepackage{subcaption}
\usepackage{threeparttable}
\usepackage{autobreak}
\usepackage{makecell}

\newcommand{\modelname}{Sequent-Prover\xspace}

\newcommand{\smtlib}{SMT-LIB\xspace}
\newcommand{\sympy}{SymPy\xspace}

\lstdefinestyle{formalstyle}{
    basicstyle=\small\ttfamily,
    keywordstyle=\color{blue},
    stringstyle=\color{red!60!black},
    commentstyle=\color{green!60!black},
    breaklines=true,
    showstringspaces=false,
    numbers=left,
    numberstyle=\tiny\color{gray},
    stepnumber=1,
    numbersep=5pt
}

\newtcblisting{leancode}[1][]{
    listing engine=listings,
    listing style=formalstyle,
    boxrule=0.5pt,
    colback=white,
    colframe=gray!70,
    arc=0pt,
    boxsep=0pt,
    left=2pt, right=2pt, top=2pt, bottom=2pt,
    listing only,
    title style={colback=white, colframe=gray!70, coltext=black},
    fonttitle=\bfseries,
    #1,
}

\newtcblisting{plainleancode}[1][]{
    listing engine=listings,
    listing only,
    listing options={
        basicstyle=\small\ttfamily,
        keywordstyle=\color{black},
        stringstyle=\color{black},
        commentstyle=\color{black},
        breaklines=true,
        showstringspaces=false,
        numbers=none
    },
    boxrule=0.5pt,
    colback=white,
    colframe=gray!70,
    arc=0pt,
    boxsep=0pt,
    left=2pt, right=2pt, top=2pt, bottom=2pt,
    breakable,
    fonttitle=\bfseries,
    #1,
}

\definecolor{commentcolor}{RGB}{0,128,0}
\definecolor{symbolcolor}{rgb}{0.5,0.2,0.8}
\definecolor{sortcolor}{rgb}{0.5,0.5,0.8}
\definecolor{darkblue}{rgb}{0, 0, 0.5}
\hypersetup{colorlinks=true, citecolor=darkblue, linkcolor=darkblue, urlcolor=darkblue}

\usepackage{colortbl}
\definecolor{gain}{RGB}{26,127,72}
\definecolor{loss}{RGB}{196,40,48}
\definecolor{oursrow}{RGB}{236,243,251}
\newcommand{\gain}[1]{\textcolor{gain}{#1}}
\newcommand{\loss}[1]{\textcolor{loss}{#1}}
\newcommand{\ours}{\rowcolor{oursrow}}
\newcommand{\passhat}[1]{\textit{pass\textasciicircum{}#1}}

\usepackage{listings}
\usepackage{xcolor}
\usepackage{booktabs}
\usepackage{hyperref}
\usepackage{enumitem}
\usepackage{microtype}

\definecolor{smtkw}{rgb}{0.2,0.2,0.7}
\definecolor{smtcomment}{rgb}{0.4,0.6,0.4}
\definecolor{smtstring}{rgb}{0.7,0.2,0.2}
\definecolor{smtbg}{gray}{0.97}

\lstdefinelanguage{SMT}{
  morekeywords={declare-const,declare-fun,declare-sort,declare-datatypes,
                define-fun,assert,check-sat,get-model,get-value,set-logic,set-option,
                push,pop,echo,Bool,Int,Real,not,and,or,xor,implies,iff,
                forall,exists,ite,distinct,let},
  sensitive=true,
  morecomment=[l]{;},
  morestring=[b]"
}
\title{{\modelname: Training Agents for \\ Formal, Checkable SMT-based Reasoning}}

\author{
  \begin{minipage}{\textwidth}
  \centering 
  Debargha Ganguly\textsuperscript{1}, Weicong Chen\textsuperscript{1},
  Mohsen Hariri\textsuperscript{1}, Vikash Singh\textsuperscript{1},
  Sreehari Sankar\textsuperscript{1}, Shouren Wang\textsuperscript{1},
  Van Yang\textsuperscript{1}, Biyao Zhang\textsuperscript{1},
  Chaoda Song\textsuperscript{1}, Alexander Nemecek\textsuperscript{1},
  Kai Ye\textsuperscript{1}, Osama Zafar\textsuperscript{1},
  Erman Ayday\textsuperscript{1}, Joe Stubbs\textsuperscript{2},
  Srinivasan Iyengar\textsuperscript{3}, Vipin Chaudhary\textsuperscript{1}\\[4pt]
  \mdseries
  \textsuperscript{1}Case Western Reserve University,
  \textsuperscript{2}UT Austin,
  \textsuperscript{3}Microsoft Research\\[1pt]
  {\small\raisebox{-0.15em}{\includegraphics[height=1em]{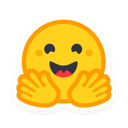}}\,%
  \texttt{Sequent-Prover-\{\href{https://huggingface.co/saccha-ai/Sequent-Prover-4B-Preview}{4B},%
  \href{https://huggingface.co/saccha-ai/Sequent-Prover-9B-Preview}{9B},%
  \href{https://huggingface.co/saccha-ai/Sequent-Prover-27B-Preview}{27B}\}}\quad
  \faGithub\,\href{https://github.com/DebarghaG/proofofthought}{\texttt{proofofthought}}}
  \end{minipage}
}

\begin{document}
\maketitle

\begin{abstract}

We introduce Sequent-Prover, a family of verifier-using language-model agents, and the post-training framework that produces them. Agents reason in an executable SMT-LIB scratchpad: they write complete SMT-LIB programs, query Z3 across turns, inspect solver feedback, repair formalizations, and terminate with an explicit answer. Our recipe, verifier-conditioned trajectory distillation, bootstraps from a noisy teacher proposal policy by sampling multi-turn trajectories, retaining the successful tail that is correct, non-trivial, grounded, and executable, then applying supervised fine-tuning and trajectory-level preference optimization. Because solver success alone does not guarantee semantic faithfulness, we evaluate task accuracy, k-sample reliability, and SMT program quality. Across 37 benchmarks and 7,254 problems per model family, post-training raises the 4B model's per-attempt accuracy from 85.6\% to 97.0\% and four-sample all-correct reliability (\passhat{4}) from 64.1\% to 91.7\%. It also reduces average SMT calls from 4.10 to 2.52 and tokens per attempt from 18.97k to 12.72k. Sequent-Prover agents give answers backed by executable formal artifacts.

\end{abstract}

\section{Introduction}

Large language models can produce fluent multi-step reasoning, but fluency is not executable evidence that a conclusion follows from the stated assumptions~\citep{wei2022chain,ZeroShot_CoT2022,turpin2023language,lanham2023measuring}. A natural alternative is to make reasoning \emph{checkable}: translate an input into explicit assumptions, constraints, and claims, then verify the resulting obligation with a symbolic backend~\citep{pan2023logiclm, satlm, olausson2023linc, ganguly2024proof}. SMT solvers such as Z3 are especially attractive because SMT-LIB can encode logical entailment, arithmetic, datatypes, arrays, bit-vectors, and program constraints, while solver execution returns concrete feedback such as \texttt{sat}, \texttt{unsat}, countermodels, parser errors, and type errors~\citep{de2008z3,barrett2018satisfiability,barrett2025smtlib}.

\begin{figure}[!htbp]
    \centering
    \includegraphics[width=0.9\linewidth]{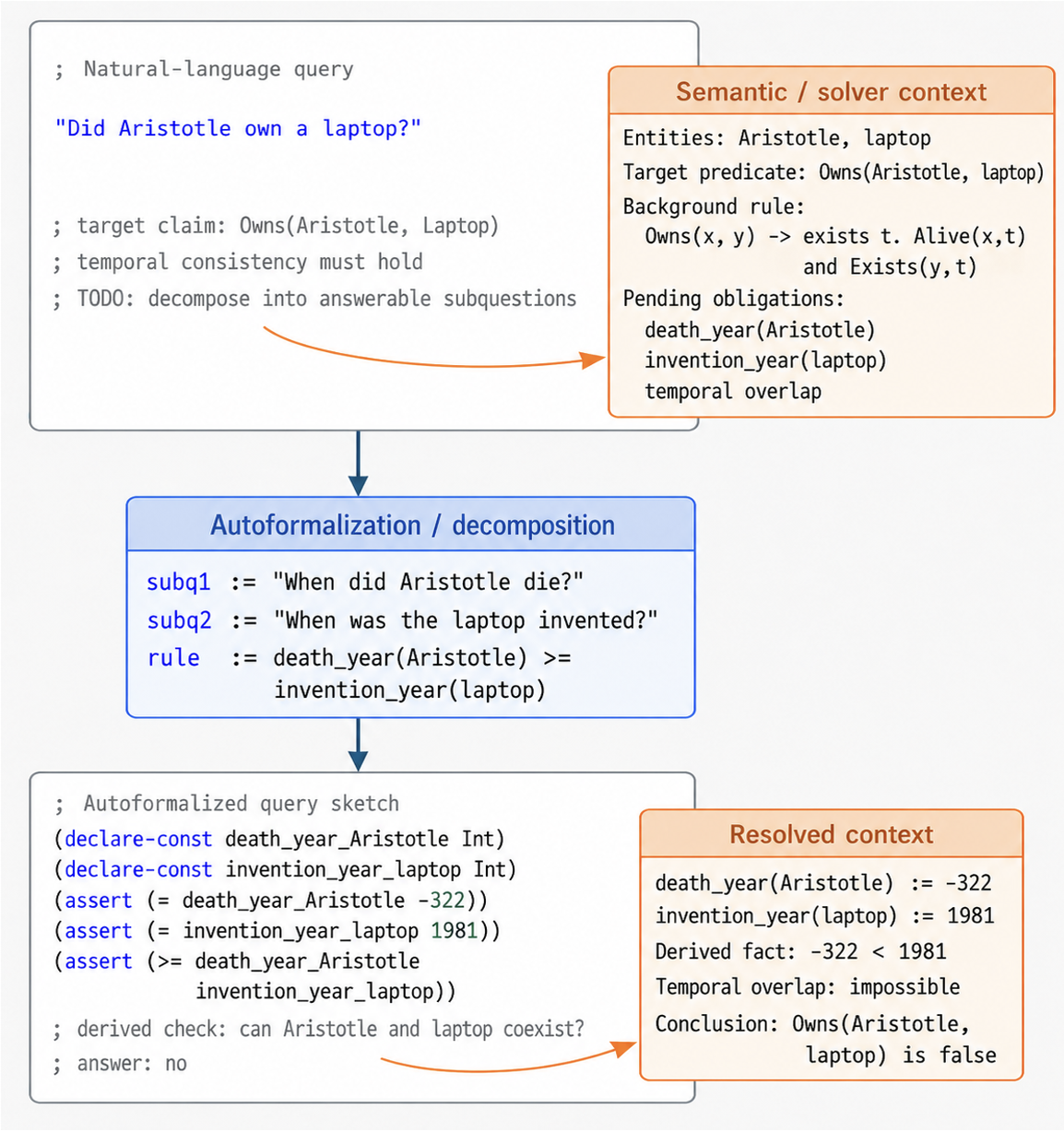}
\caption{A toy example showing how Sequent-Prover turns an underspecified commonsense claim into a grounded, checkable inconsistency.}
    \label{fig:toy}

\end{figure}
However, solver access alone does not make reasoning reliable. Recent work by~\citet{ganguly2026grammars} shows that SMT-based autoformalization can improve premise-explicit reasoning but degrade performance when the generated artifact misrepresents the input. The core problem is the \emph{autoformalization gap}: real tasks arrive as natural language, specifications, and informal intent, not as already-correct proof obligations. A solver checks only the artifact it receives; if the model omits a premise, reverses entailment, introduces inconsistent assumptions, or encodes a different problem, the solver may return a valid result for the wrong object. Reliable verifier use therefore requires models that construct grounded, non-vacuous, executable formalizations through multi-turn tool use.

We introduce \emph{Sequent-Prover}, a family of post-trained language models for SMT-verifiable reasoning. Rather than using the solver as a test-time wrapper~\citep{ganguly2024proof, singh2026verge, feng2025vericot}, we use solver interaction as a training substrate: we sample candidate formalization traces, execute them with Z3, reject malformed or vacuous artifacts, and convert supported traces into supervised and preference data. Post-training turns inconsistent SMT tool use into more dependable multi-sample correctness and yields agents that write more grounded, non-trivial, executable SMT programs. In particular, we find that:

\begin{enumerate}[nosep,leftmargin=1.3em]
    \item Post-training substantially improves SMT reasoning: models that previously achieved only inconsistent solver-aided success become more reliable under multi-sample evaluation.

    \item Supervised fine-tuning improves \emph{proof discipline}: tuned agents produce higher-quality SMT-LIB programs that are better grounded in the input, less vacuous, and structurally more consistent across samples.

    \item Post-training strengthens more demanding proof strategies, including refutational reasoning, where agents increasingly formulate proof obligations by contradiction and correctly interpret the resulting \texttt{unsat} evidence.

    \item Preference optimization with DPO further improves stability, grounding, proof quality, and efficiency beyond supervised fine-tuning: DPO-tuned agents solve problems earlier, with fewer tokens and fewer solver-interaction rounds per correct answer.
\end{enumerate}

\begin{figure*}[!h]
    \centering
    \includegraphics[width=1\linewidth]{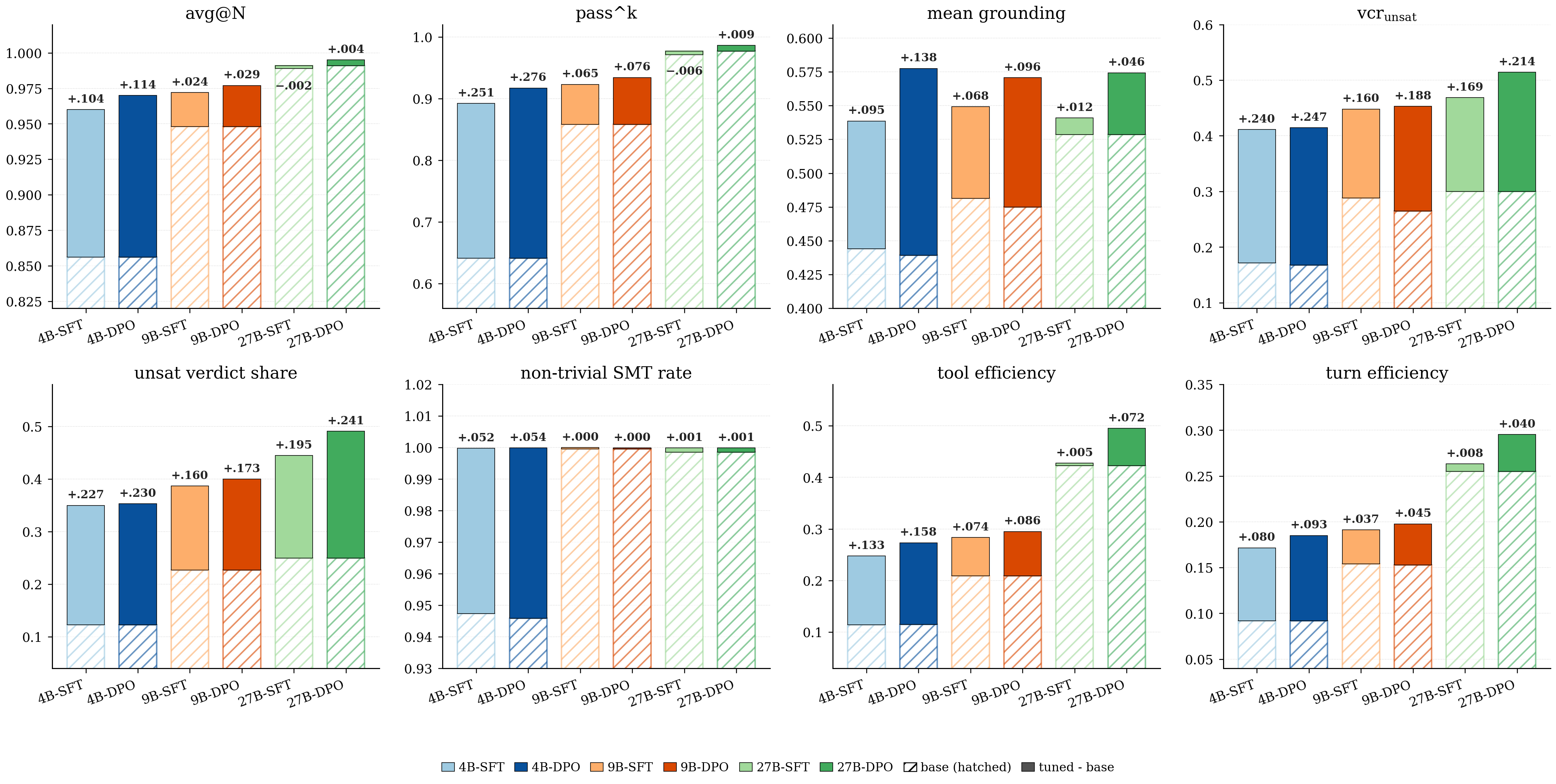}
    \caption{ Post-training, especially DPO, shifts the models from merely sampling correct answers toward more reliable solver-grounded proof behavior. Hatched bars show each base checkpoint and colored overlays show the tuned-model gain: avg@N measures per-sample correctness, \passhat{k} measures consistency by requiring all sampled trajectories to be correct, mean grounding measures how much of the problem is represented in SMT, and vcr\_unsat measures correct answers backed by non-trivial UNSAT/refutation certificates. The remaining panels show the mechanism behind the gains: tuned models more often drive Z3 to UNSAT verdicts, avoid trivial SMT calls, and obtain more verified progress per tool call and per reasoning turn.}
    \label{fig:gains}
    \vspace{-1em}
\end{figure*}

\begin{figure*}
        \centering
        \includegraphics[width=1\linewidth]{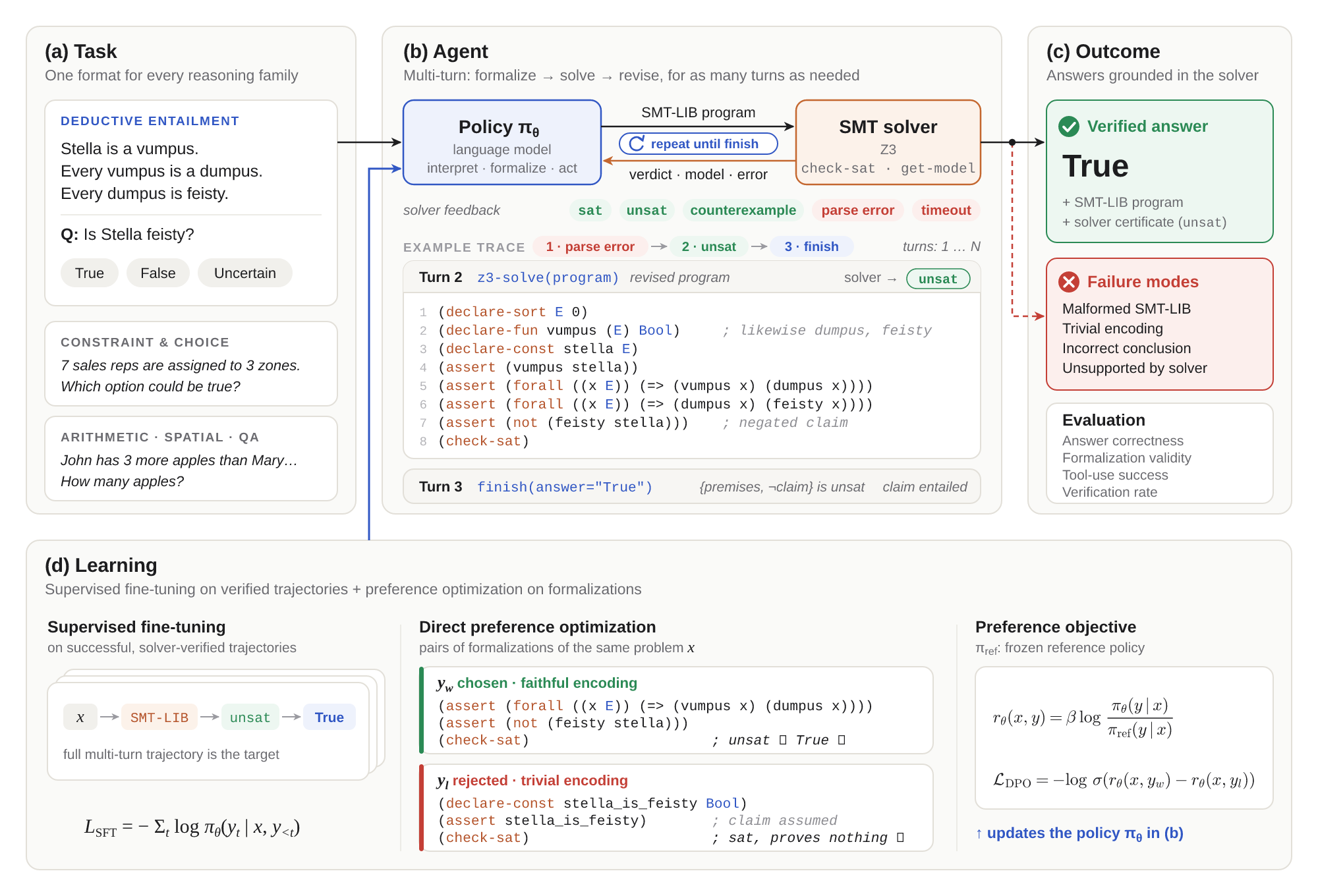}
        \caption{Sequent-Prover overview. The agent formalizes a task in SMT-LIB, calls Z3 over several turns, and uses solver feedback until it finishes. Successful trajectories train SFT, and preferred versus rejected trajectories train DPO.}
        \vspace{-1em}
        \label{fig:framework}
\end{figure*}

\section{Methodology}
\label{sec:method}

\paragraph{Overview.} We frame verifier-guided reasoning as success-conditioned policy bootstrapping. A teacher model need not be reliable on every rollout; it only needs a nonzero probability on trajectories that use the verifier correctly. Sampling the teacher in an SMT-LIB scratchpad exposes a sparse successful tail: rollouts that construct and revise executable formalizations, use Z3 feedback, and terminate with the correct answer. We condition on this tail so the student learns verifier-mediated search behavior that the teacher produces only sporadically.

Our method, \emph{verifier-conditioned trajectory distillation}, has three stages. First, we sample teacher trajectories under the same \texttt{z3-solve}/\texttt{finish} interface used at evaluation time. Second, we curate them with an acceptance predicate that combines answer correctness with executable scratchpad evidence: solver execution, non-triviality, grounding, and protocol completion. Third, we train by supervised fine-tuning on accepted traces and by DPO on trajectory-level pairs that contrast accepted traces with rejected or weakly supported ones. The resulting policy amortizes rare successful verifier-mediated searches into more reliable inference.

\subsection{Problem Formulation}
Each task instance is $x=(\mathcal{C},q)$, where $\mathcal{C}=\{p_1,\ldots,p_m\}$ contains the natural-language premises, program fragments, or specifications, and $q$ is the query. The model must produce a final answer $y$ matching $y^\star$ and, when possible, support it with an executable SMT-LIB artifact. We use $\phi$ for the model's formalization map: $\phi_{\mathcal{C}}$ encodes assumptions from $\mathcal{C}$, and $\phi_c$ encodes a candidate claim $c$. Context consistency is checked by $\mathrm{SAT}(\phi_{\mathcal{C}})$; claim compatibility by $\mathrm{SAT}(\phi_{\mathcal{C}}\wedge \phi_c)$; and entailment by refutation, requiring $\mathrm{SAT}(\phi_{\mathcal{C}})=\mathrm{true}$ and $\mathrm{SAT}(\phi_{\mathcal{C}}\wedge \neg \phi_c)=\mathrm{false}$. Thus, a claim is entailed only when the formalized context is satisfiable and the claim's negation is inconsistent with it, making \texttt{unsat} the proof signal for refutational entailment.

\subsection{The SMT Scratchpad Interface}

A central design choice is to expose the SMT-LIB backend from \citet{ganguly2024proof} as an \emph{executable scratchpad}: a workspace in which the model constructs, tests, and revises formal artifacts across turns. We define the tool environment as $\mathcal{E}_{\mathrm{SMT}}=(\mathcal{A},\mathcal{O}), \mathcal{A}=\{\texttt{z3-solve},\texttt{finish}\}.$ The \texttt{z3-solve} action executes a self-contained SMT-LIB script with declarations and a \texttt{check-sat} command. A single action can express context checks, candidate-claim tests, counterexample searches, model construction, semantic-equivalence checks, or refutational proof obligations. Z3 returns an observation in $\mathcal{O}$, such as \texttt{sat}, \texttt{unsat}, \texttt{unknown}, a model, a parser/type error, or a timeout, which is appended to the interaction history.

The \texttt{finish} action is separate from scratchpad execution: it terminates the episode with a final answer after the model judges that the accumulated solver feedback is sufficient. A trajectory is therefore a multi-turn record $\tau=(m_1,o_1,\ldots,m_T,o_T),$ where model messages may contain natural-language reasoning, SMT-LIB code, tool calls, repairs, or the final answer, and the policy $\pi_\theta$ maps the current history to the next token or tool action. We do not require every scratchpad call to constitute a complete proof; early calls may be exploratory or diagnostic. However, accepted training traces must contain meaningful scratchpad evidence: at least one Z3 invocation with an executable, non-trivial, input-grounded SMT-LIB program that obtains a real solver verdict.

\subsection{Verifier Feedback and Trace Semantics}

Each \texttt{z3-solve} call converts a generated SMT-LIB artifact into a structured observation that the agent can use to interpret, repair, or commit to a formalization. We write $\mathrm{SAT}(\psi)\in\{\texttt{sat}, \texttt{unsat},\texttt{unknown}\}$ for the Z3 verdict on a formula or script $\psi$, and treat parser errors, type errors, timeouts, malformed scripts, and solver failures as unresolved observations. For a candidate claim $c_i$ with context encoding $\phi_{\mathcal{C}}$ and claim encoding $\phi_{c_i}$, the verifier first separates inconsistent contexts from meaningful claim statuses: if $\mathrm{SAT}(\phi_{\mathcal{C}})=\texttt{unsat}$, the artifact receives the inconsistent-context status $\sigma_I$. Otherwise, when $\mathrm{SAT}(\phi_{\mathcal{C}})=\texttt{sat}$, the claim is entailed ($\sigma_E$) if $\mathrm{SAT}(\phi_{\mathcal{C}}\wedge\neg \phi_{c_i})=\texttt{unsat}$, contradictory ($\sigma_C$) if $\mathrm{SAT}(\phi_{\mathcal{C}}\wedge\phi_{c_i})=\texttt{unsat}$, and possible but not forced ($\sigma_P$) if both $\mathrm{SAT}(\phi_{\mathcal{C}}\wedge\phi_{c_i})=\texttt{sat}$ and $\mathrm{SAT}(\phi_{\mathcal{C}}\wedge\neg\phi_{c_i})=\texttt{sat}$. We use $\sigma_U$ for states in which no semantic status can be assigned, including \texttt{unknown}, timeout, solver failure, or parse/type errors.  The same scratchpad also supports comparison between alternative formalizations: given formulas $\phi_a$ and $\phi_b$ over a shared signature and background theory $\Sigma$, they are equivalent when $\mathrm{SAT}_{\Sigma}(\neg(\phi_a\leftrightarrow\phi_b))=\texttt{unsat}$. 

Solver feedback is richer than a binary success signal: \texttt{sat} can expose a witness or countermodel, \texttt{unsat} provides refutational evidence, and execution errors identify artifacts that must be repaired. When constraints are labeled or instrumented, this feedback can be augmented with unsat cores or approximate correction sets, but the main training signal comes from executable SMT-LIB interaction and solver feedback.

\paragraph{Teacher Rollout Generation.} For each training instance $x=(\mathcal{C},q)$, we sample $K_T$ trajectories from a teacher proposal model $M_T$ under the same SMT-scratchpad interface used at evaluation time. The teacher is given only the problem instance $x$, not the target answer $y^\star$; the target is reserved for offline curation. Thus, the teacher is not an oracle annotator, but a stochastic proposal distribution $p_T(\tau\mid x)$ over executable reasoning traces. Each sampled trajectory is stored as $\tau=(x,h_T,\hat{y},\mu), h_T=(m_1,o_1,\ldots,m_T,o_T), $ where $h_T$ is the interaction transcript, $\hat{y}$ is the answer emitted through \texttt{finish}, and $\mu$ records protocol metadata such as solver-call counts, solver verdicts, execution errors, token usage, and termination status. These rollouts provide the candidate pool from which verifier-conditioned trace curation selects training examples.

\paragraph{Verifier-Conditioned Trace Curation.} The offline curator converts teacher rollouts into training data by applying a trajectory-level acceptance predicate \(A(x,y^\star,\tau)\). A rollout is accepted when it reaches the correct task answer through \texttt{finish} and contains meaningful scratchpad evidence: an executable \texttt{z3-solve} interaction that obtains a solver verdict and passes the structural checks for non-triviality, grounding, and detectable non-vacuity. We label accepted rollouts as \emph{Gold}. Rollouts that reach the correct answer but provide weak formal support are labeled \emph{Silver}, while rollouts that fail the answer, protocol, executability, or scratchpad-quality conditions are labeled \emph{Rejected}. This partition separates task success from verifier-use quality, so that only fully supported traces become demonstrations while weak or failed traces remain available as a contrastive signal.

Formally, if \(p_T(\tau\mid x)\) is the teacher proposal distribution and \(Z_T(x)=\Pr_{\tau\sim p_T(\cdot\mid x)}[A(x,y^\star,\tau)=1]\), then the gold supervision distribution is $
p_{\mathrm{gold}}(\tau\mid x)
=
\frac{p_T(\tau\mid x)A(x,y^\star,\tau)}{Z_T(x)}
=
p_T(\tau\mid x, A(x,y^\star,\tau)=1).
$
Trace curation defines the success-conditioned target that SFT
imitates, with Silver and Rejected traces supplying the negative side of the
preference data.

\subsection{Training}

\paragraph{SFT on Executable Traces.} Gold trajectories define the supervised corpus
$\mathcal{D}_{\mathrm{gold}}$. We train the student policy $\pi_\theta$ by
maximum likelihood over the model-generated actions in each accepted
transcript:
\[
\mathcal{L}_{\mathrm{SFT}}(\theta)
=
-
\sum_{\tau \in \mathcal{D}_{\mathrm{gold}}}
\sum_{t \in \mathcal{I}(\tau)}
\log \pi_\theta(a_t \mid h_t),
\]
where $h_t$ is the interaction history, $a_t$ is the next model token or
tool-use action, and $\mathcal{I}(\tau)$ indexes positions generated by the
model. Solver observations remain in the conditioning history but are masked
from the prediction loss. SFT amortizes the curated successful tail: the
student learns the action patterns by which accepted traces construct evidence,
respond to verifier feedback, and commit to a correct answer.

\paragraph{Trajectory Preference Optimization.}
After SFT, we further refine the student with trajectory-level preferences.
For each problem \(x\) with at least one Gold trace and at least one
contrastive trace, we form triples \((x,\tau^+,\tau^-)\), where
\(\tau^+\) is an accepted trajectory and \(\tau^-\) is either Rejected or
Silver. Silver traces serve as hard negatives because they may reach the
correct answer with weaker scratchpad support; Rejected traces capture answer,
protocol, executability, or verifier-use failures. Gold traces define
what to favor, while Silver and Rejected traces define nearby behaviors to
suppress.

We optimize Direct Preference Optimization with the frozen SFT model
\(\pi_{\mathrm{ref}}\) as reference. Let
\(r_\theta(\tau,x)=\log \pi_\theta(\tau\mid x)-
\log \pi_{\mathrm{ref}}(\tau\mid x)\). The objective is
\(
\mathcal{L}_{\mathrm{DPO}}(\theta)
=
-\mathbb{E}_{(x,\tau^+,\tau^-)}
\log \sigma\!\left(
\beta\left[
r_\theta(\tau^+,x)-r_\theta(\tau^-,x)
\right]
\right),
\)
where \(\beta\) controls the strength of the preference update. As in SFT,
trajectory likelihoods are computed over model-generated tokens and tool
actions, with solver observations used as conditioning context. This stage
raises the probability of complete, verifier-supported traces and lowers the probability of correct-looking but weakly supported or failed ones.

\paragraph{Policy Bootstrapping.} The success-conditioned bootstrapping mechanism relies on the teacher's \emph{support} under the acceptance predicate. Let
\(Z_T(x)=\Pr_{\tau\sim p_T(\cdot\mid x)}[A(x,y^\star,\tau)=1]\) denote the
teacher's probability of producing an accepted trajectory for problem \(x\).
When \(Z_T(x)>0\), sampling \(K_T\) rollouts exposes at least one Gold trace
with probability \(1-(1-Z_T(x))^{K_T}\). Trace curation turns these rare
successes into a high-precision target: SFT amortizes accepted search behavior, while DPO uses Silver and Rejected traces to suppress nearby failure modes.

\paragraph{Inference.} At test time, the student uses the same SMT-scratchpad protocol as teacher sampling and training. Given \(x=(\mathcal{C},q)\), it generates tool-use actions until \texttt{finish} or the interaction budget. We score the final answer against \(y^\star\) and retain the full trajectory for process analysis: solver invocation, verdict quality, non-triviality, grounding, and protocol completion. For multi-sample evaluation, we draw \(K\) independent trajectories per problem and aggregate answer and scratchpad-quality metrics.

\providecommand{\base}{\textsc{Base}}
\providecommand{\sft}{\textsc{SFT}}
\providecommand{\dpo}{\textsc{DPO}}
\providecommand{\best}[1]{\textbf{#1}}

\section{Results}
\label{sec:results}

\paragraph{Evaluation protocol.}
We present evaluations of nine Sequent-Prover checkpoints: base, supervised fine-tuned
(\sft), and further preference-optimized (\dpo) variants of the Qwen3.5--4B,
Qwen3.5--9B, and Qwen3.6--27B families.  Each model is evaluated with
$K=4$ stochastic SMT-scratchpad trajectories per problem.  The final paired
pool contains 37 benchmarks and 7,254 problems per model family.  The suite
covers logical and constraint reasoning (ProofWriter, RuleTaker, FOLIO,
ProntoQA, AR-LSAT, ZebraLogic, ReClor, StrategyQA), arithmetic and spatial
reasoning (GSM8K, SVAMP, MAWPS, ASDiv, AQuA-RAT, DROP, SpartQA-MC,
SpartQA-YN, StepGame), NLI and multiple-choice commonsense/knowledge tasks
(ANLI, BoolQ, WinoGrande, CommonsenseQA, OpenBookQA, QASC, ARC-Challenge,
MMLU, HellaSwag), domain and reading-comprehension QA (MedQA, MedMCQA,
PolicyQA, HotpotQA, MuSiQue, SQuAD, TriviaQA), and AGIEval LSAT/SAT subsets
(LSAT-LR, LSAT-RC, SAT-Math, SAT-EN).

\paragraph{Metrics.}
We report \emph{avg@$N$}, the per-attempt accuracy over all sampled
trajectories; \emph{pass@4}, the probability that at least one of four
samples solves the problem; and \passhat{4}, the probability that all
four samples solve the problem.  We use \passhat{4} as the main reliability
metric because deployed stochastic agents should produce correct behavior
consistently, rather than only under best-of-$K$ selection.  We also report
\emph{stability}=\passhat{4}/pass@4, which normalizes all-sample consistency by
four-sample coverage.  For process analysis, we use the SMT-quality metrics
from Section~\ref{sec:method}: first-pass Z3 validity, non-trivial grounded
solver use, solver verdict distributions, and token/iteration efficiency.

\paragraph{Research questions.}
The results address four questions.  \textbf{RQ1}: does verifier-conditioned
post-training improve single-sample accuracy and multi-sample reliability,
beyond what extra test-time sampling gives?  \textbf{RQ2}: how do \sft{} and
\dpo{} differ, and which parts of the recipe matter?  \textbf{RQ3}:
how does post-training change the solver-agent process?  \textbf{RQ4}: where
are the gains largest across tasks and model scales?

\begin{table*}[!h]
\centering
\footnotesize
\setlength{\tabcolsep}{3.2pt}

\begin{minipage}[t]{0.64\textwidth}
\centering
\textbf{Aggregate performance}\\[-0.00em]
\begin{tabular}{@{}llrrrrr@{}}
\toprule
\textbf{Family} & \textbf{Ckpt.} & \textbf{avg@$N$ $\uparrow$} &
\textbf{pass@4 $\uparrow$} & \textbf{\passhat{4} $\uparrow$} &
\textbf{Stab. $\uparrow$} & \textbf{$\Delta$\passhat{4}} \\
\midrule
4B & \base & 0.856 & 0.970 & 0.641 & 0.661 & -- \\
4B & \sft  & 0.960 & 0.995 & 0.892 & 0.896 & \gain{+0.251} \\
4B & \dpo  & \best{0.970} & \best{0.998} & \best{0.917} & \best{0.919} & \gain{\best{+0.277}} \\
\midrule
9B & \base & 0.948 & 0.994 & 0.858 & 0.863 & -- \\
9B & \sft  & 0.972 & 0.997 & 0.923 & 0.926 & \gain{+0.065} \\
9B & \dpo  & \best{0.977} & \best{0.998} & \best{0.934} & \best{0.936} & \gain{\best{+0.076}} \\
\midrule
27B & \base & 0.991 & 0.998 & 0.977 & 0.979 & -- \\
27B & \sft  & 0.989 & 0.999 & 0.971 & 0.972 & \loss{$-$0.006} \\
27B & \dpo  & \best{0.995} & \best{1.000} & \best{0.986} & \best{0.986} & \gain{\best{+0.009}} \\
\bottomrule
\end{tabular}
\end{minipage}
\hfill
\begin{minipage}[t]{0.35\textwidth}
\centering
\textbf{pass@$k$ scaling}\\[-0.00em]
\begin{tabular}{@{}lrrrr@{}}
\toprule
\textbf{Model} & \textbf{@1} & \textbf{@2} & \textbf{@3} & \textbf{@4} \\
\midrule
4B \base & 0.856 & 0.943 & 0.963 & 0.970 \\
&&&& \\
4B \dpo  & \best{0.970} & \best{0.992} & \best{0.996} & \best{0.998} \\
\midrule
9B \base & 0.948 & 0.984 & 0.991 & 0.994 \\
&&&& \\
9B \dpo  & \best{0.977} & \best{0.994} & \best{0.997} & \best{0.998} \\
\midrule
27B \base & 0.991 & 0.996 & 0.998 & 0.998 \\
&&&& \\
27B \dpo  & \best{0.995} & \best{0.999} & \best{1.000} & \best{1.000} \\
\bottomrule
\end{tabular}
\end{minipage}

\caption{Aggregate performance and pass@$k$ scaling on the 37-benchmark paired evaluation pool. 
Each family is compared against its own base checkpoint. pass@4 measures four-sample coverage, while \passhat{4} requires all four sampled trajectories to be correct. 
The pass@$k$ panel shows that a single 4B \dpo{} sample matches the 4B base model's four-sample coverage. Bold marks the best checkpoint within each family; $\Delta$ is relative to \base{}, with \gain{gains} and \loss{losses} colored.}
\vspace{-1em}
\label{tab:main-results}
\end{table*}

\subsection{Aggregate accuracy and reliability}
\label{sec:results:aggregate}

Table~\ref{tab:main-results} shows the main comparison.  Verifier-conditioned
post-training yields the largest improvements for the 4B model: \dpo{} raises
avg@$N$ from 0.856 to 0.970 and \passhat{4} from 0.641 to 0.917, a 27.7-point
increase in all-sample reliability.  The 9B model follows the same pattern
from a stronger starting point, with \dpo{} improving \passhat{4} from 0.858 to
0.934.  The 27B base model is already close to saturation on the aggregate
pool, and \dpo{} still reaches the best value in the family for every reported
metric, including \passhat{4} of 0.986.

The distinction between pass@4 and \passhat{4} captures the central effect.  The
base checkpoints already have high four-sample coverage, especially at 9B and
27B, due to the multi-option nature of most of these tasks.  Post-training concentrates probability mass on the correct trajectory mode, producing large gains in \passhat{4} and stability.  For 4B, stability
rises from 0.661 to 0.919 under \dpo; for 9B, it rises from 0.863 to 0.936.
This shows that the trained policies improve single-run reliability as well
as best-of-four coverage.

Table~\ref{tab:main-results} presents the pass@$k$ curves for the base and \dpo{}
checkpoints.  The 4B result is especially clear: 4B \dpo{} reaches 0.970 at
pass@1, matching the 4B base model's pass@4 value.  At 9B, \dpo{} raises
pass@1 from 0.948 to 0.977 and nearly saturates pass@4.  At 27B, the curve is
near the ceiling throughout, with \dpo{} reaching 1.000 by pass@3.

\begin{table*}[!t]
\centering
\scriptsize
\setlength{\tabcolsep}{2.5pt}
\renewcommand{\arraystretch}{0.86}

\begin{subtable}[t]{0.57\textwidth}
\centering
\caption{Problem-level transitions grouped by base-model correctness.}
\label{tab:transitions}
\resizebox{\linewidth}{!}{%
\begin{tabular}{@{}llrrrrrr@{}}
\toprule
\textbf{Family} & \textbf{Tune} & \textbf{$n_{0/4}$} &
\textbf{$P(\geq1/4\mid0/4)$} & \textbf{$n_{1-3/4}$} &
\textbf{$P(4/4\mid1$--$3/4)$} & \textbf{$n_{4/4}$} &
\textbf{$P(4/4\mid4/4)$} \\
\midrule
4B  & \sft & 215 & 0.930 & 2,379 & 0.839 & 4,660 & 0.942 \\
4B  & \dpo & 215 & \best{0.958} & 2,379 & \best{0.867} & 4,660 & \best{0.965} \\
\midrule
9B  & \sft & 41 & \best{0.780} & 970 & 0.753 & 6,243 & 0.954 \\
9B  & \dpo & 41 & \best{0.780} & 970 & \best{0.787} & 6,243 & \best{0.962} \\
\midrule
27B & \sft & 13 & 0.846 & 153 & 0.536 & 7,088 & 0.982 \\
27B & \dpo & 13 & \best{0.923} & 153 & \best{0.719} & 7,088 & \best{0.993} \\
\bottomrule
\end{tabular}%
}
\end{subtable}
\hfill
\begin{subtable}[t]{0.4\textwidth}
\centering
\caption{Direct \dpo{}--\sft{} comparison. Deltas are macro-averages over shared datasets; DPO-only/SFT-only paired solves are 25/8, 13/8, and 7/0 for 4B, 9B, and 27B.}
\label{tab:dpo-vs-sft}

\resizebox{\linewidth}{!}{%
\begin{tabular}{@{}lrrrrrrr@{}}
\toprule
\textbf{Family} & \textbf{$\Delta$avg@$N$} & \textbf{$\Delta$\passhat{4}} &
\textbf{$\Delta$stab.} & \textbf{$\Delta$VCR} &
\textbf{$\Delta$VCR$_{\textsc{unsat}}$} \\
\midrule
4B  & +0.005 & +0.009 & +0.007 & +0.050 & +0.013 \\
9B  & +0.006 & +0.008 & +0.011 & +0.036 & +0.026\\
27B & +0.005 & +0.012 & +0.016 & +0.046 & +0.050  \\
\bottomrule
\end{tabular}%
}
\end{subtable}

\vspace{-0.5em}
\caption{Performance transition and direct-comparison analyses for tuned checkpoints. 
In Table~\ref{tab:transitions}, the three rates measure coverage gained on base-unsolved problems, conversion of partially correct problems to 4/4 correctness, and retention of already 4/4 problems. 
In Table~\ref{tab:dpo-vs-sft}, metric deltas are macro-averages over shared datasets; DPO-only and SFT-only count paired problems solved by one tuned checkpoint and not the other.}
\label{tab:transition-dpo-summary}
\vspace{-1em}
\end{table*}

\subsection{Problem-level consistency transitions}
\label{sec:results:transitions}

Table~\ref{tab:transitions} decomposes the reliability gains by the number of
correct base-model attempts.  Across 4B and 9B, the largest source of
improvement is the middle tier: problems with one to three correct base
attempts become correct on all four tuned attempts.  For 4B \dpo, this happens
for 86.7\% of the 2,379 partially correct base problems.  For 9B \dpo, it
happens for 78.7\% of the 970 partially correct base problems.  The tuned
models also preserve most problems that were already reliable: \dpo{} retains
96.5\%, 96.2\%, and 99.3\% of base 4/4 problems in the 4B, 9B, and 27B
families, respectively.

The transition table also clarifies the role of scale.  The 4B base model has
2,379 partially correct problems, so post-training has substantial headroom to
improve reliability.  The 27B base model has only 153 partially correct
problems and 13 zero-correct problems, so aggregate gains are naturally smaller.
Within that restricted headroom, \dpo{} still improves the middle-tier
conversion rate from 53.6\% under \sft{} to 71.9\%, and it preserves nearly
all already reliable examples.

\vspace{-0.75em}
\subsection{Training-stage ablation}
\label{sec:results:ablation}

Table~\ref{tab:dpo-vs-sft} compares \dpo{} directly against \sft{} on shared
datasets.  Preference optimization improves every family on avg@$N$, \passhat{4},
stability, and verified-correct rate.  The coverage overlap is large: for
example, 4B \sft{} and 4B \dpo{} both solve 7,213 problems at least once,
while \dpo{} adds 25 DPO-only problems and \sft{} adds 8 SFT-only problems.
The most consistent DPO gains are in reliability and trace quality.
At 27B, the DPO stage is particularly useful: it adds +0.012 macro \passhat{4},
+0.016 stability, +0.046 verified-correct rate, and +0.050
UNSAT-restricted verified-correct rate over SFT.

\begin{table}[t]
\centering
\small
\setlength{\tabcolsep}{3.5pt}
\begin{tabular}{@{}lrrrr@{}}
\toprule
\textbf{Method} & \textbf{$K$} & \textbf{pass@$K$ $\uparrow$} & \textbf{\passhat{K} $\uparrow$} & \textbf{Tokens $\downarrow$} \\
\midrule
Base               & 4  & 0.848 & 0.395 & 110.9k \\
Base               & 16 & 0.911 & 0.139 & 443.7k \\
Base + self-refine & 4  & 0.928 & 0.572 & 219.5k \\
\midrule
\ours Post-trained 4B    & 1  & 0.869 & 0.869 & \best{16.7k} \\
\ours Post-trained 4B    & 4  & \best{0.948} & \best{0.743} & 66.9k \\
\bottomrule
\end{tabular}
\caption{Post-training versus test-time sampling (4B, 180 held-out problems from six families). Tokens are per problem. Shaded rows are post-trained. Full results are in Table~\ref{tab:ttc}.}
\vspace{-1em}
\label{tab:ttc-main}
\end{table}

\begin{table}[t]
\centering
\small
\setlength{\tabcolsep}{3.5pt}
\resizebox{\linewidth}{!}{%
\begin{tabular}{@{}lrrrr@{}}
\toprule
\textbf{Checkpoint} & \textbf{avg@$N$ $\uparrow$} & \textbf{pass@4 $\uparrow$} & \textbf{\passhat{4} $\uparrow$} & \textbf{Stab. $\uparrow$} \\
\midrule
\ours SFT (ours)             & 0.585 & 0.715 & 0.409 & 0.572 \\
\ours DPO (ours)             & \best{0.603} & \best{0.718} & 0.439 & 0.611 \\
\midrule
Answer-only SFT        & 0.407 & 0.654 & \loss{0.126} & 0.193 \\
Solver-verdict SFT     & 0.583 & 0.712 & 0.403 & 0.566 \\
Continued SFT          & 0.591 & 0.700 & \best{0.440} & \best{0.629} \\
Gold/Silver/Rej.\ DPO  & 0.586 & 0.699 & 0.419 & 0.599 \\
\bottomrule
\end{tabular}}
\caption{Data-selection ablations for 4B on five hard held-out benchmarks.}
\vspace{-1em}
\label{tab:data-ablations}
\end{table}

\subsection{Test-time sampling and data ablations}
\label{sec:results:ttc}

\textbf{Post-training versus more samples.}
We compare the 4B models on 180 held-out problems, 30 each from ProntoQA,
FOLIO, AR-LSAT, GSM8K, StepGame, and ARC-Challenge, with the same harness,
temperature, and iteration limit for every method (Appendix~\ref{app:add:ttc}).  At matched $K=4$,
post-training raises pass@4 by 10.1 points and \passhat{4}
by 34.8 points over base self-consistency, with fewer tokens
(Table~\ref{tab:ttc-main}).  The post-trained model at $K=4$ also beats the
base model at $K=16$ on both metrics while using 6.6$\times$ fewer tokens.
From $K=1$ to $K=16$, base pass@$K$ rises from 0.657 to 0.911, but
\passhat{K} falls from 0.657 to 0.139.  Extra sampling
raises coverage but lowers all-sample reliability.
Self-refinement, the strongest prompting baseline, still trails the
post-trained model while using 3.3$\times$ more tokens.

\textbf{Which parts of the recipe matter.}
We train four 4B variants and evaluate them on ReClor, SpartQA-MC,
SpartQA-YN, StepGame, and AGIEval-LSAT-LR (200 problems each, $K=4$;
Table~\ref{tab:data-ablations}).  \emph{Answer-only SFT} keeps only the final answer, and
\passhat{4} falls from 0.409 (full-trajectory SFT) to 0.126, so the gains do not come
from label exposure.  \emph{Solver-verdict SFT} removes the grounding and
non-triviality checks and is lower on every metric.  \emph{Continued SFT}
replaces DPO with more SFT at matched training FLOPs.  It ties DPO on
\passhat{4} and is slightly higher on stability, but lower on
avg@$N$ and pass@4.
\emph{Gold/Silver/Rejected DPO} uses the three tiers as separate preference
levels and is below our DPO on every metric.
Our full recipe leads on avg@$N$ and pass@4.

\begin{table*}[!t]
\centering
\scriptsize
\renewcommand{\arraystretch}{0.90}

\newcommand{\theadx}[1]{%
\begin{tabular}[c]{@{}c@{}}#1\end{tabular}%
\rule{0pt}{3.2ex}\rule[-1.4ex]{0pt}{0pt}}

\begin{subtable}[t]{0.48\textwidth}
\vspace{0pt}
\centering
\caption{Solver-agent efficiency.}
\label{tab:efficiency}

\setlength{\tabcolsep}{2.2pt}

\begin{tabular*}{\linewidth}{@{\extracolsep{\fill}}llrrrrr@{}}
\toprule
\textbf{Family} & \textbf{Ckpt.} &
\textbf{Iter. $\downarrow$} &
\textbf{\theadx{SMT\\calls $\downarrow$}} &
\textbf{Tokens $\downarrow$} &
\textbf{\theadx{Tok./\\correct $\downarrow$}} &
\textbf{\theadx{First VC\\iter. $\downarrow$}} \\
\midrule
4B & \base & 4.97 & 4.10 & 18.97k & 26.73k & 2.00 \\
4B & \sft  & 3.76 & 2.78 & 13.70k & 15.23k & 1.33 \\
4B & \dpo  & \best{3.51} & \best{2.52} & \best{12.72k} & \best{13.89k} & \best{1.29} \\
\midrule
9B & \base & 4.37 & 3.41 & 15.69k & 19.53k & 1.62 \\
9B & \sft  & 3.52 & 2.54 & 12.31k & 14.16k & 1.24 \\
9B & \dpo  & \best{3.33} & \best{2.34} & \best{11.85k} & \best{13.27k} & \best{1.22} \\
\midrule
27B & \base & 2.92 & 1.90 & 9.51k & 10.28k & 1.19 \\
27B & \sft  & 2.85 & 1.86 & 7.92k & 8.52k & 1.08 \\
27B & \dpo  & \best{2.58} & \best{1.58} & \best{6.78k} & \best{7.01k} & \best{1.07} \\
\bottomrule
\end{tabular*}
\end{subtable}
\hfill
\begin{subtable}[t]{0.48\textwidth}
\vspace{0pt}
\centering
\caption{SMT scratchpad diagnostics.}
\label{tab:smt-quality}

\setlength{\tabcolsep}{2.2pt}

\begin{tabular*}{\linewidth}{@{\extracolsep{\fill}}llrrrrr@{}}
\toprule
\textbf{Family} & \textbf{Ckpt.} &
\textbf{Error $\downarrow$} &
\textbf{\theadx{No\\verdict $\downarrow$}} &
\textbf{UNSAT $\uparrow$} &
\textbf{\theadx{UNSAT\\\textrm{|}correct $\uparrow$}} &
\textbf{Calls} \\
\midrule
4B & \base & 41.3\% & 10.3\% & 12.3\% & 12.9\% & 122.6k \\
4B & \sft  & 19.1\% & \best{0.3\%} & 35.0\% & 36.2\% & 82.0k \\
4B & \dpo  & \best{19.0\%} & 0.4\% & \best{35.3\%} & \best{36.3\%} & 73.3k \\
\midrule
9B & \base & 26.0\% & 0.4\% & 22.7\% & 23.0\% & 101.1k \\
9B & \sft  & \best{13.9\%} & \best{0.3\%} & 38.7\% & 39.7\% & 75.0k \\
9B & \dpo  & 14.4\% & 0.4\% & \best{40.0\%} & \best{41.0\%} & 68.7k \\
\midrule
27B & \base & 16.9\% & 0.7\% & 25.0\% & 25.2\% & 59.8k \\
27B & \sft  & 5.8\% & \best{0.1\%} & 44.5\% & 45.8\% & 55.0k \\
27B & \dpo  & \best{5.5\%} & \best{0.1\%} & \best{49.1\%} & \best{49.7\%} & 46.9k \\
\bottomrule
\end{tabular*}
\end{subtable}

\vspace{-0.45em}

\caption{Efficiency and SMT-scratchpad diagnostics.
In Table~\ref{tab:efficiency}, Iter. is the average number of model iterations, SMT calls is the average number of solver invocations, Tokens is the average token count per problem attempt, Tok./correct normalizes token use by correct-attempt rate, and First VC iter. is the mean iteration index of the first verified-correct SMT signal.
In Table~\ref{tab:smt-quality}, Error and no-verdict measure unusable calls, UNSAT is the share of calls returning an UNSAT verdict, UNSAT$\mid$correct restricts to calls inside correct attempts, and Calls is total solver calls.}
\vspace{-1em}
\label{tab:efficiency-smt-quality}

\end{table*}

\subsection{Solver-agent process quality}
\label{sec:results:process}

Table~\ref{tab:efficiency} shows that post-training improves the computational
profile of the agent loop.  The 4B \dpo{} checkpoint reduces average
iterations from 4.97 to 3.51, SMT calls from 4.10 to 2.52, and tokens from
18.97k to 12.72k.  Normalizing by successful attempts, tokens per correct
answer fall from 26.73k to 13.89k.  The 9B and 27B families show the same
pattern, with \dpo{} giving the lowest token and solver-call cost within each
family.

The first verified-correct SMT signal also arrives earlier after tuning.  The
mean first-success iteration drops from 2.00 to 1.29 for 4B \dpo, from 1.62 to
1.22 for 9B \dpo, and from 1.19 to 1.07 for 27B \dpo.  Under a two-iteration
budget, 4B \dpo{} reaches a verified-correct rate of 0.902, above the 4B base
model's full-budget rate of 0.765; 9B \dpo{} reaches 0.929 by iteration two,
above the 9B base full-budget rate of 0.908.  These early-exit curves show
that post-training improves both endpoint accuracy and the efficiency frontier
of the interactive solver loop.

Table~\ref{tab:smt-quality} shows that the tuned policies emit cleaner SMT
programs and obtain stronger solver signals.  For 4B, the error rate falls
from 41.3\% to 19.0\% under \dpo, and no-verdict calls fall from 10.3\% to
0.4\%.  For 27B, the error rate falls from 16.9\% to 5.5\%.  The same table
shows a shift toward refutational solver evidence: UNSAT calls increase from
12.3\% to 35.3\% for 4B \dpo, from 22.7\% to 40.0\% for 9B \dpo, and from
25.0\% to 49.1\% for 27B \dpo.  The UNSAT share inside correct attempts rises
in parallel, reaching 49.7\% for 27B \dpo.  The grounding gains hold with SMT
comments removed (+0.081 for 4B \dpo; Appendix~\ref{app:add:grounding}).
Post-training also lowers the structural (PCFG) divergence between independent
SMT programs in all six Base-to-tuned comparisons, with every 95\% interval
below zero (Appendix~\ref{app:add:pcfg}).  Such structural variation predicts
autoformalization errors~\citep{ganguly2026grammars}.

\begin{table}[t]
\centering
\small
\setlength{\tabcolsep}{2.0pt}
\begin{tabular}{@{}llrrrr@{}}
\toprule
\textbf{Fam} & \textbf{Dataset} & \textbf{Base \passhat{4}} &
\textbf{DPO \passhat{4}} & \textbf{$\Delta$} \\
\midrule
4B & ProofWriter & 0.074 & 0.926 & \gain{+0.852}  \\
4B & ZebraLogic & 0.120 & 0.715 & \gain{+0.595}\\
4B & RuleTaker & 0.395 & 0.985 & \gain{+0.590} \\
4B & AR-LSAT & 0.035 & 0.560 & \gain{+0.525} \\
4B & ProntoQA & 0.445 & 0.955 & \gain{+0.510}\\
4B & SpartQA-MC & 0.250 & 0.745 & \gain{+0.495}\\
\midrule
9B & AR-LSAT & 0.235 & 0.640 & \gain{+0.405}\\
9B & ZebraLogic & 0.410 & 0.790 & \gain{+0.380} \\
9B & QASC & 0.840 & 1.000 & \gain{+0.160} \\
9B & WinoGrande & 0.845 & 0.990 & \gain{+0.145} \\
9B & CommonsenseQA & 0.845 & 0.980 & \gain{+0.135} \\
9B & ARC-Challenge & 0.870 & 0.985 & \gain{+0.115}\\
\midrule
27B & ZebraLogic & 0.825 & 0.965 & \gain{+0.140} \\
27B & AR-LSAT & 0.785 & 0.920 & \gain{+0.135} \\
27B & StepGame & 0.825 & 0.860 & \gain{+0.035} \\
27B & ReClor & 0.960 & 0.985 & \gain{+0.025}  \\
\bottomrule
\end{tabular}
\caption{Largest representative dataset-level \dpo{} gains on \passhat{4} within each model family. }
\vspace{-2em}
\label{tab:dataset-gains}
\end{table}

\subsection{Benchmark-level trends}
\label{sec:results:benchmarks}

Table~\ref{tab:dataset-gains} reports representative per-dataset improvements
for \dpo.  The highest-gain tasks are those with explicit symbolic or
relational structure.  At 4B, \passhat{4} improves by 85.2 points on ProofWriter,
59.0 points on RuleTaker, 52.5 points on AR-LSAT, 59.5 points on ZebraLogic,
and 51.0 points on ProntoQA.  At 9B, the largest gains include AR-LSAT,
ZebraLogic, QASC, WinoGrande, CommonsenseQA, and ARC-Challenge.  At 27B,
aggregate scores are close to ceiling, and the clearest remaining gains are on
ZebraLogic and AR-LSAT.  On AR-LSAT, all three \dpo{} students exceed the
teacher's \passhat{3} of 0.484, reaching 0.560, 0.640, and
0.920 (Appendix~\ref{app:add:teacher}).

These dataset-level results align with the scratchpad interpretation of the
method.  The strongest gains occur where the task contains latent constraints,
relational structure, arithmetic structure, or entailment structure that can be
made executable.  Broad QA and domain tasks also benefit in the smaller model
families, with improvements on ARC-Challenge, CommonsenseQA, MedQA, MedMCQA,
OpenBookQA, QASC, and WinoGrande.  The aggregate effect spans more
than formal logic benchmarks while remaining largest on tasks with natural SMT
structure.  At 4B and 9B,
\dpo{} improves avg@$N$ and \passhat{4} in every category,
and hard reasoning is only 14.5\% of the pool
(Appendix~\ref{app:add:categories}).

\begin{table}[!htbp]
\centering
\small
\setlength{\tabcolsep}{3.6pt}
\begin{tabular}{@{}lrrrr@{}}
\toprule
\textbf{Comparison} & \textbf{Small} & \textbf{Large} & \textbf{Tie} & \textbf{Win $\uparrow$} \\
\midrule
4B \sft{} / 9B \base & 816 & 518 & 5,920 & 0.612 \\
4B \dpo{} / 9B \base & 840 & 361 & 6,053 & \best{0.699} \\
9B \sft{} / 27B \base & 73 & 490 & 6,691 & 0.130 \\
9B \dpo{} / 27B \base & 90 & 420 & 6,744 & \best{0.176} \\
4B \dpo{} / 27B \base & 84 & 532 & 6,638 & 0.136 \\
\bottomrule
\end{tabular}
\caption{Cross-scale paired comparison.  Small and Large count paired problems
where the smaller tuned checkpoint or larger base checkpoint has more correct
attempts out of four.  Win is the smaller tuned checkpoint's non-tie win rate.}
\vspace{-1em}
\label{tab:scale}
\end{table}

\textbf{Post-training and scale.}
\label{sec:results:scale} Table~\ref{tab:scale} compares tuned smaller models against larger base
models.  The 4B tuned agents compare favorably with the 9B base model: 4B
\sft{} wins 61.2\% of non-tied paired comparisons, and 4B \dpo{} wins 69.9\%.
The 27B base model sets a higher aggregate ceiling, and the tuned 9B models win
fewer non-tied comparisons against it.  Combining scale and post-training gives
the strongest overall checkpoints: 27B \dpo{} has the highest aggregate
avg@$N$, pass@4, \passhat{4}, and SMT-quality profile in the sweep.

\section{Related Works}\label{sec:related}

\paragraph{Verifier-Integrated Formal Reasoning.}
Recent progress in formal theorem proving has shown that language models can benefit greatly from emitting machine-checkable artifacts rather than only natural-language explanations. Systems such as Goedel Prover, DeepSeek-Prover, and Kimina-Prover combine long-form reasoning with formal verification and iterative refinement \citep{lin2025goedel,ren2025deepseek,wang2025kimina}. Our work is inspired by this paradigm, but shifts the target language from Lean to \smtlib and focuses on compact models that must operate across multiple SMT logics.

\paragraph{Solver-Based Symbolic Reasoning and Synthesis.}
SMT solvers such as Z3 and cvc5 are central tools for formal verification, constraint solving, and program synthesis \citep{de2008z3,barbosa2022cvc5}. Classical symbolic systems rely on hand-written encodings or search procedures over formal objects. Our setting is different: the primary generator is a language model trained to emit solver-executable artifacts directly, while a symbolic backend is used to validate outputs and to synthesize additional training data. The use of \sympy as a controllable symbolic front-end is especially important for scaling supervision across diverse logic fragments \citep{meurer2017sympy}.

\paragraph{Self-Repair and Tool Feedback.}
Outside theorem proving, a large body of work studies how models can improve outputs by reacting to execution feedback, compiler messages, or explicit critics \citep{yao2023tree,shinn2023reflexion,first2023baldur,olausson2024is,chen2024teaching,bouzenia2024repairagent}. We adopt the same high-level intuition, but the repair channel here is solver-native: parser errors, sort mismatches, failed assertions, and counterexamples. This makes the feedback more structured than free-form critique and better aligned with symbolic correctness.

\paragraph{Reinforcement Learning for Verifiable Outputs.}
RL has recently become an effective way to improve pass@1 performance in reasoning and formal generation tasks \citep{shao2024deepseekmath,liu2025understanding,yu2025dapo}. Our work extends that lesson to SMT-native generation. Because repair is serial and strongly depends on first-round quality, improving the expected solver reward is particularly useful in our setting.

\section{Conclusion}
\label{sec:conclusion}

Sequent-Prover shows that solver use can be learned as a formal-reasoning policy. By filtering trajectories with executable SMT feedback and contrasting successful traces against failed ones, our method amortizes solver-guided search into the model: it learns to formulate proof obligations, encode them in SMT-LIB, interpret Z3 responses, repair formal artifacts, and commit with evidence. Empirically, post-training turns sample-dependent solver successes into stable behavior, produces more grounded and non-vacuous SMT programs, and shifts correct traces toward refutational \textsc{unsat} evidence. The scale-dependent results show where this training helps most: when a base model has latent solver-use competence but lacks consistency. Training on solver interaction is a direct way to make language-model reasoning checkable.

\section{Limitations}

Sequent-Prover inherits a basic limit of formal verification: a solver can check a formal artifact, but it cannot decide whether the chosen abstraction is the intended one. Expert verification engineers face the same problem when they decide what to model, what to omit, which predicates or theories to use, and which specification captures the informal task. Sequent-Prover makes this step executable and learnable: it trains models to build grounded, non-trivial SMT artifacts, use solver feedback, and revise failed abstractions. Most tasks have no reference formalization, and semantic equivalence is co-NP-complete in propositional logic and undecidable in first-order logic. No automatic check can certify meaning preservation, so our grounding, non-triviality, and structural-uncertainty measures (Appendix~\ref{app:add:pcfg}) serve as strong proxies for it. Learning the right level of formal abstraction remains an open problem.

\bibliography{references}

\appendix

\section{Background: Agent, Solver \& Training}
\label{app:background}

We first describe three pieces of background: (1) \emph{what Z3 actually computes}, (2) \emph{how the language model interacts with it}, and (3) \emph{how the two post-training recipes differ in what they teach}. The metric definitions that follow depend on these choices.

\subsection{The Z3 SMT Solver and the Two Proof Modes}
\label{app:bg:z3}

Z3 is a Satisfiability Modulo Theories (SMT) solver: given a quantifier-free first-order formula $\varphi$ over a combination of background theories (linear/non-linear integer or real arithmetic, fixed-size bit-vectors, equality with uninterpreted functions, strings,
arrays, datatypes), Z3 returns one of
\begin{itemize}
  \setlength{\itemsep}{0.1em}
  \item \textsc{sat}: $\varphi$ has a model under the theory
    combination, and Z3 (optionally) returns it as
    $(\texttt{define-fun }x\ldots)$ definitions;
  \item \textsc{unsat}: $\varphi$ is unsatisfiable in the
    theory combination, so no assignment can make it true;
  \item \textsc{unknown}: Z3's procedure was incomplete or
    timed out;
  \item an \textsc{error} string: the input did not parse, used a
    symbol Z3 cannot interpret, or violated the declared sort.
\end{itemize}
Our agent uses Z3 in two operationally distinct modes; the distinction is central to how to read the verdict mix in \S\ref{app:metrics:smt}.

\paragraph{Direct constraint-satisfaction (the \textsc{sat} mode).} The agent encodes the problem's premises as constraints and the candidate answer as an additional assertion, then queries \texttt{(check-sat)}. A return of \textsc{sat} establishes that the candidate answer is \emph{consistent} with the premises and the agent may extract a satisfying model from Z3's \texttt{(get-model)} output. For an open-ended constraint-search task (e.g., ZebraLogic, where the agent must find ``the house with the cat''), the witness \emph{is} the answer. For a deductive task (e.g., ProofWriter, where the agent must say whether ``$p$ entails $q$''), a \textsc{sat} verdict on $\Gamma \cup \{q\}$ only tells us that $q$ is \emph{not} contradicted
by $\Gamma$, which is a strictly weaker claim than $\Gamma \models q$.

\paragraph{Proof by contradiction (the \textsc{unsat} mode).} The deductively-strong way to use Z3 for an entailment question is to assert the premises together with the \emph{negation} of the candidate conclusion, $\Gamma \cup \{\neg q\}$, and check satisfiability. A return of \textsc{unsat} establishes that $\Gamma \cup \{\neg q\}$ has no model, i.e.\ $\Gamma \models q$ in the theory. This is the standard refutational argument that underlies essentially all SMT-based theorem proving. An agent that systematically produces \textsc{unsat} verdicts on deductive tasks is using the solver as a \emph{prover}; an agent that systematically produces \textsc{sat} verdicts on the same tasks is using it as a \emph{model-finder} and is, in the worst case, verifying nothing more than internal consistency.

The pair of metrics $\mathrm{vcr}$ and $\mathrm{vcr}_{\textsc{unsat}}$
introduced in \S\ref{app:metrics:smt} make this distinction
quantitative: a delta in $\mathrm{vcr}_{\textsc{unsat}}$ larger than
the delta in $\mathrm{vcr}$ indicates that post-training is shifting
the agent toward refutational rather than model-finding use of the
solver, even when the underlying correctness count is unchanged.

\subsection{The Tool-Calling Agent Loop}
\label{app:bg:agent}

The agent harness uses an \emph{OpenAI-compatible} tool-calling
protocol exposed by vLLM. At each iteration $\tau$ the model
receives the cumulative conversation $\mathcal{H}_\tau =
\bigl(m_1, m_2, \ldots\bigr)$ consisting of the system prompt, the
user problem, every prior model turn, and the structured payload
from every prior tool call. The model may emit one of two output
shapes:
\begin{enumerate}
  \setlength{\itemsep}{0.1em}
  \item a \emph{tool call} to \texttt{z3\_solve} with arguments
    $\{\texttt{smt\_code} : s_\tau,\, \texttt{timeout} : \texttt{int}\}$,
    parsed by Qwen's \texttt{qwen3\_coder} tool grammar; or
  \item a \emph{free-text} reply containing the final answer.
\end{enumerate}
Reasoning content (Qwen's $\langle$think$\rangle$\ldots$\langle$/think$\rangle$
block) is parsed by the \texttt{qwen3} reasoning parser and is not
re-emitted to subsequent turns; only the model's user-visible reply
plus its tool-call schema is carried forward, which prevents the
context window from growing quadratically in $T_{\max}$.

When the agent calls \texttt{z3\_solve}, the harness runs $s_\tau$
under a 30-second wall-clock cap and a 256-MB memory cap, captures
the full stdout/stderr, and returns the payload $(\texttt{success},
\texttt{output}, \texttt{sat\_result}, \texttt{error})$ as a
tool-response message. The \texttt{sat\_result} field is taken from
the exit line of \texttt{(check-sat)} and normalised to one of
$\{$\textsc{sat}, \textsc{unsat}, \textsc{unknown}$\}$. The original
output is kept separately so the agent (and our analyzer) can recover
$(\texttt{get-model})$ witnesses or error stacks.

The loop terminates when the model produces an attempt without a
tool call (success), when it explicitly emits the special
\texttt{finish} terminator (success), or when $\tau$ reaches
$T_{\max} = 10$ (timeout). The terminating attempt's free-text
answer is run through the dataset-specific \texttt{check\_correct}
extractor (which handles option-letter matching for MCQ,
multiset matching for ZebraLogic, exact-string for arithmetic, etc.)
to produce the correctness bit $c_p^{(k)}$.

The boolean attribute \texttt{verified} on each JSONL row is
\emph{not} the correctness bit. It records whether the agent
terminated by producing an answer (\texttt{verified}~=~\textsc{True})
or by hitting the iteration cap and being forced to extract whatever
the model last said (\texttt{verified}~=~\textsc{False}); the metrics
\emph{soundness} and \emph{completeness} in \S\ref{app:metrics:agent}
treat \texttt{verified} as a confidence/commitment signal, not as
correctness.

\subsection{Datasets.}
We sample 1{,}000 questions each from the training splits of three hard reasoning datasets: AR-LSAT~\citep{zhong2021arlsat}, FOLIO~\citep{han2024folio}, and ZebraLogic~\citep{lin2025zebralogic}. For diversity, we add a small hand-curated set from BBEH~\citep{kazemi2025bbeh}, CASP~\citep{hertzberg2025casp}, Code2Inv~\citep{si2018code2inv}, DafnyBench~\citep{loughridge2024dafnybench}, and SV-COMP~\citep{beyer2023svcomp}. The three core datasets were chosen because they require different forms of symbolic reasoning: AR-LSAT consists of analytical reasoning problems with combinatorial constraints, FOLIO requires reasoning over first-order logical premises, and ZebraLogic contains constraint-satisfaction puzzles with latent entity--attribute assignments. This gives the model supervision on problems where success typically requires constructing and checking an explicit symbolic state, rather than relying only on lexical or parametric associations.

At evaluation time, we enforce strict decontamination: no problem used for training is included in evaluation, and evaluation examples are drawn from held-out splits whenever labels are available. We therefore distinguish between \emph{in-distribution} evaluation, which consists only of held-out examples from the same three core dataset families used during training, and \emph{out-of-distribution} evaluation, which covers all other benchmark sources. This distinction is at the dataset-family level rather than the instance level: even the in-distribution setting is instance-disjoint from training, while the out-of-distribution setting tests transfer to new annotation protocols, answer formats, domains, and reasoning styles.

The evaluation suite is broad. In addition to the three core training benchmarks, we evaluate on logical and rule-based reasoning tasks such as StrategyQA, ProntoQA, ProofWriter, ReClor, and RuleTaker; arithmetic and discrete reasoning benchmarks such as GSM8K, SVAMP, MultiArith, ASDiv, AQuA-RAT, MAWPS, and DROP; spatial reasoning tasks including SpartQA and StepGame; commonsense and science multiple-choice benchmarks including OpenBookQA, QASC, ARC-Challenge, CommonsenseQA, and WinoGrande; broad-coverage multiple-choice benchmarks including MMLU, HellaSwag, BoolQ, MedQA, and MedMCQA; and free-form multi-hop QA datasets including HotpotQA, MuSiQue, SQuAD, and TriviaQA
~\citep{
geva2021strategyqa,
saparov2022language,
tafjord2021proofwriter,
han2024folio,
zhong2024agieval, lin2025zebralogic, yu2020reclor, clark2020transformers, cobbe2021training, patel2021nlp,
roy2015solving, miao2020diverse, ling2017program, koncel2016mawps, dua2019drop, mirzaee2021spartqa, shi2022stepgame,mihaylov2018can, khot2020qasc, clark2018think, talmor2019commonsenseqa, sakaguchi2021winogrande, hendrycks2020measuring, zellers2019hellaswag, clark2019boolq, yang2018hotpotqa, trivedi2022musique, rajpurkar2016squad, joshi2017triviaqa}. The resulting split tests whether post-training on a small set of hard symbolic datasets teaches a general solver-use behavior that transfers beyond the observed training distribution, rather than merely improving accuracy on the benchmark families used for supervision.

\subsection{Reproducibility.} We have released the SMT harness and its code (\url{https://github.com/DebarghaG/proofofthought}). We have also released the post-trained 4B, 9B, and 27B models (\url{https://huggingface.co/saccha-ai}).

\textbf{Training Data.} Our trajectory dataset contains a total of 14{,}307 training samples, containing 10{,}785 gold, 2{,}995 reject, 464 silver, and 63 discarded infrastructure-error trajectories. This data was generated by using our verifier-conditioned trajectory distillation from a Deepseek-v4 Flash model. Temperature during both dataset curation and all evaluation was maintained at 0.9.

\textbf{Teacher rollouts and acceptance.} We sample three teacher rollouts per training problem (eight for part of AR-LSAT). Table~\ref{tab:teacher-acceptance} gives the teacher's answer accuracy on the three core datasets. A trace is Gold only when the final answer matches the label, the trace ends with \texttt{finish}, the SMT is non-trivial and grounded, and Z3 returns a real verdict. Labels are checked with the same dataset-specific \texttt{check\_correct} extractor used at evaluation. Rejected traces are incorrect, unfinished, malformed, or have no SMT call. On AR-LSAT, many rejected traces keep calling \texttt{z3-solve} without finishing; others contain real formalization errors. A few correct answers were rejected because they did not match the exact output format. On FOLIO, failures are more semantic: the teacher translates premises into predicates inconsistently and mixes up True, False, and Uncertain. We de-duplicate training and evaluation problems, so no training problem appears in the test pool.

\begin{table}[h]
\centering
\small
\begin{tabular}{@{}lrr@{}}
\toprule
\textbf{Dataset} & \textbf{pass@3} & \textbf{\passhat{3}} \\
\midrule
AR-LSAT    & 0.727 & 0.484 \\
FOLIO      & 0.853 & 0.738 \\
ZebraLogic & 0.999 & 0.901 \\
\bottomrule
\end{tabular}
\caption{Teacher (DeepSeek-v4 Flash) answer accuracy with three rollouts per training problem.}
\label{tab:teacher-acceptance}
\end{table}

\paragraph{Training details.}
We post-train models using full-parameter fine-tuning with DeepSpeed ZeRO-3. For supervised fine-tuning, we optimize the autoregressive language-modeling objective over pre-rendered Qwen3.5 chat-template traces. Each training example is stored as a single text sequence that preserves the exact interleaving of reasoning, tool calls, and solver outputs, including \texttt{<think>...</think>}, \texttt{<tool\_call>...</tool\_call>}, and \texttt{<tool\_response>...</tool\_response>} spans. We use this representation directly rather than converting through an OpenAI- or ShareGPT-style message format, since the latter can discard assistant reasoning text when tool calls are also present. This ensures that the model is trained on the same rendered traces that it is expected to produce at inference time.

The supervised run uses the base checkpoint, full fine-tuning, bfloat16 precision, gradient checkpointing, FlashAttention, and a maximum sequence length of 16{,}384 tokens. We enable sequence packing for throughput. Training is performed for two epochs on 8$\times$A100-80GB GPUs or H200 GPUs according to SLURM availability with per-device batch size 1 and gradient accumulation 8, giving an effective batch size of 64. We use AdamW with learning rate $3\times10^{-6}$, cosine learning-rate decay, warmup ratio 0.03, weight decay 0, and gradient clipping at norm 1.0. Checkpoints are saved every 10 optimizer steps, retaining the three most recent checkpoints.

We then apply Direct Preference Optimization (DPO) starting from the last supervised checkpoint, where the supervised loss had plateaued in our run. The preference data consists of chosen/rejected SMT autoformalization traces. We train on the entire assistant trajectory, while preserving the same Qwen3.5 chat-template rendering. We use the standard sigmoid DPO objective with $\beta=0.1$. DPO is also performed with full-parameter fine-tuning, bfloat16 precision, gradient checkpointing, FlashAttention, and DeepSpeed ZeRO-3. Because DPO evaluates chosen and rejected continuations separately and does not use packing, we set the cutoff length to 8{,}192 tokens and discard preference pairs whose prompt-plus-response length exceeds this limit, avoiding truncation inside tool-call spans. The DPO run is trained for two epochs on 8$\times$A100-80GB GPUs with per-device batch size 1 and gradient accumulation 4, giving an effective batch size of 32. We use learning rate $5\times10^{-7}$, cosine decay, warmup ratio 0.1, weight decay 0, and gradient clipping at norm 1.0.

\textbf{Scope, Design Choices and Limitations}
We note six scope decisions. First, Sequent-Prover builds on the SMT-LIB backend of Proof-of-Thought \citep{ganguly2024proof}, and our contribution is the post-training recipe. We do not compare against Grammars of Formal Uncertainty or VERGE \citep{singh2026verge, ganguly2026grammars}, which sample hundreds of programs per question, since a fair comparison would need a similar test-time budget. Section~\ref{sec:results:ttc} instead compares our 4B model against its base model with up to sixteen samples and against three prompting baselines. Second, each component ablation needs a full post-training and evaluation sweep, so we do not ablate every design choice at every scale. We ablate the main data-selection choices at 4B on five held-out benchmarks (Section~\ref{sec:results:ttc}). Across nine checkpoints and 37 benchmarks, the full recipe improves reliability, grounding, and refutational solver use, and tuning individual components may give further gains. Third, the largest improvements are on the smaller models and on tasks with latent symbolic structure. The 27B Qwen3.6 base model is already a strong tool-caller and code generator, and post-training still brings it near ceiling on the aggregate pool. Most remaining headroom lies in tasks harder than the current suite. Fourth, we report most metrics in aggregate over 7,254 problems per family. This keeps the paper readable, and the pool is large enough for stable deltas. Appendix~\ref{app:add:categories} gives per-category results for all three families. The auxiliary measures in Appendix~\ref{app:metrics}, such as soundness, recovery, calibration, and skeleton diversity, follow the same trends as the main reliability and verifier-use metrics. The predicted SFT-induced drop in soundness does not lower reliability (Appendix~\ref{app:add:soundness}). Fifth, curation and evaluation share the same grounding and triviality predicates. These permissive, substring-based criteria apply equally to every checkpoint, including the untrained baselines, so if anything they favor the baselines. The grounding trends also hold when SMT comments are removed (Appendix~\ref{app:add:grounding}). Finally, all base models and the DeepSeek-v4-Flash teacher use their own chat-template format, and the mixed point-release versions within a family are the actual base models we used. We have released the SMT harness and the post-trained checkpoints, and will release the 14,307 curated trajectories and prompts, so that the results can be reproduced with the same harness.

\textbf{License, Compute Budget, Risks} All evaluation benchmarks are public datasets used under their original research licenses; Z3 (MIT), and the Qwen base models (under their respective open model licenses) are used in accordance with their terms, and our harness and checkpoints are released, and our trajectories will be released, under a permissive open-source license for research use. Across prototyping, repeated iteration, teacher rollout generation, SFT, DPO for all three model families, the full sweep used approximately 1000 GPU-hours on A100-80GB/H200 hardware. Our work studies formal reasoning over public academic benchmarks and introduces no new data, deployment surface, or generative capability that poses foreseeable risk of harm.

\section{Evaluation Metrics}
\label{app:metrics}

Let $\mathcal{D}$ be a dataset with $M = |\mathcal{D}|$ problems, each problem $x \in \mathcal{D}$ having a gold answer $y_x^\star$. For each problem the model produces $K$ i.i.d.\ samples; the $r$-th attempt yields a trace $\tau_x^{(r)}$ and a final answer $\hat{y}_x^{(r)}$, plus a sequence of SMT calls
\[
\bigl(z_x^{(r,1)}, \ldots, z_x^{(r,S_x^{(r)})}\bigr)
\]
with associated Z3 responses
\[
\bigl(\rho_x^{(r,1)}, \ldots, \rho_x^{(r,S_x^{(r)})}\bigr).
\]
We write
\[
\chi_x^{(r)}
:=
\mathbb{1}\!\left[
\texttt{check correct}\bigl(\hat{y}_x^{(r)}, y_x^\star\bigr)
\right]
\]
\[
\chi_x
:=
\sum_{r=1}^{K} \chi_x^{(r)}
\in \{0,1,\ldots,K\}.
\]
$\chi_x$ is the number of correct attempts on problem $x$, and
$\bar{\chi}_x := \chi_x/K$ is the per-problem accuracy. The boolean
$\kappa_x^{(r)}$ denotes the commitment event, i.e.\ the
\texttt{verified} JSONL field of \S\ref{app:bg:agent}; $T_x^{(r)}$
denotes the iteration count; and $S_x^{(r)}$ denotes the number of
SMT calls. Unless stated otherwise we use $K=4$.

We use $r$ for the sample index and reserve $n$ for the sample budget
inside metrics such as pass@$n$ and maj@$n$. We group metrics into
three families: \emph{sample-based correctness}
(\S\ref{app:metrics:correctness}), which depend only on the bag of
correctness indicators $\{\chi_x^{(r)}\}$; \emph{agent-reliability
measures} (\S\ref{app:metrics:agent}), which characterize the
relationship between the agent's internal commitment signal and its
actual correctness; and \emph{SMT-program quality}
(\S\ref{app:metrics:smt}), which inspect the structural and semantic
properties of the SMT-LIB code that the agent emitted, independent of
whether the final answer was correct.

\subsection{Sample-Based Correctness}
\label{app:metrics:correctness}

The metrics in this subsection depend only on the multiset
$\{\chi_x^{(r)}\}_{x,r}$; they do not inspect the agent trace. Each is
a moment of the per-problem correctness count
$\chi_x \in \{0,\ldots,K\}$ under a different functional form, designed
to isolate different facets of the underlying success distribution.

\paragraph{avg@$N$.}
The per-attempt expected accuracy, computed as the empirical mean of
$\bar{\chi}_x$ across problems:
\begin{equation}
\textrm{avg@}N
=
\frac{1}{M}\sum_{x \in \mathcal{D}} \bar{\chi}_x
=
\frac{1}{MK}\sum_{x,r} \chi_x^{(r)}.
\label{eq:avgN}
\end{equation}
$\textrm{avg@}N$ has the operational interpretation of
\emph{single-attempt success}: if a user runs the agent once with one
temperature draw, $\textrm{avg@}N$ is the expected correctness of that
single attempt. It is invariant to which of the $K$ samples we examine,
and it is an unbiased estimate of the per-attempt correctness
probability.

\paragraph{pass@$n$.}
The unbiased estimator of \citet{chen2021evaluating} for the probability
that at least one of $n$ samples is correct: for $n \leq K$,
\begin{equation}
\textrm{pass@}n
=
\frac{1}{M}\sum_{x \in \mathcal{D}}
\left[
1 -
\frac{\binom{K-\chi_x}{n}}{\binom{K}{n}}
\right].
\label{eq:passk}
\end{equation}
The estimator is unbiased under random subsampling without replacement
from the $K$ observed attempts: $\binom{K-\chi_x}{n}/\binom{K}{n}$ is
the exact probability of drawing $n$ \emph{all-incorrect} attempts, and
$\textrm{pass@}n$ is one minus its expectation. $\textrm{pass@}1 =
\textrm{avg@}N$ exactly. For $n>1$, the quantity reflects the user-facing
utility of taking multiple samples and accepting the answer if any one is
correct. This is an \emph{ensemble-with-oracle-selection} ceiling. We report
$\textrm{pass@}4$ throughout.

\paragraph{\passhat{n}.}
The expected probability that \emph{every} one of $n$ samples is correct,
under the same random-subsampling estimator:
\begin{equation}
\textit{pass}^n
=
\frac{1}{M}\sum_{x \in \mathcal{D}}
\frac{\binom{\chi_x}{n}}{\binom{K}{n}}.
\label{eq:passhatk}
\end{equation}
$\binom{\chi_x}{n}/\binom{K}{n}$ is the probability that all $n$
randomly drawn attempts are correct. Two checkpoints with the same
$\textrm{pass@}n$ may differ dramatically in $\textit{pass}^n$:
$\textit{pass}^n$ rewards \emph{distributional concentration on the
right answer}: high values arise iff the right answer is the modal
answer across samples, low values arise when correctness is
sample-dependent.

\paragraph{maj@$n$.}
The expected probability that majority vote among $n$ samples agrees with
the gold answer. Because \texttt{check correct} is binary, majority on
correctness is well-defined and gives, under random subsampling without
replacement and the conservative tie convention of strict majority, with
ties broken against the agent,
\begin{equation}
\begin{split}
\textrm{maj@}n
&=
\frac{1}{M}\sum_{x \in \mathcal{D}}
\frac{1}{\binom{K}{n}}
\sum_{j=\lfloor n/2 \rfloor + 1}^{n}
\binom{\chi_x}{j}
\\
&\qquad\qquad{}\times
\binom{K-\chi_x}{n-j}.
\end{split}
\label{eq:majk}
\end{equation}
We report $\textrm{maj@}4$ throughout. The gap between $\textrm{maj@}n$
and $\textrm{avg@}N$ measures the benefit of self-consistency
aggregation, which assumes wrong answers are more diverse than right ones.

\paragraph{Stability.}
The ratio of consistency to coverage,
\begin{equation}
\textrm{stability}
=
\frac{\textit{pass}^n}{\textrm{pass@}n}
\in [0,1].
\label{eq:stability}
\end{equation}
A value of $1$ indicates perfect consistency: any problem the model can
solve, it solves on \emph{every} attempt. A value of $0$ indicates the
right answer is purely sample-dependent. The stability delta summarizes
the main post-training effect in one number.

\paragraph{AUC@$n$.}
A finer-grained ``area-under-the-curve'' summary of the full
$(\textrm{pass@}j)_{j=1}^{n}$ trajectory, used internally by the
reporting pipeline:
\begin{equation}
\textrm{AUC@}n
=
\frac{1}{n}\sum_{j=1}^{n} \textrm{pass@}j.
\end{equation}
$\textrm{AUC@}n$ favors checkpoints whose accuracy at small $j$ is
already high, i.e., checkpoints that do not depend on the
oracle-selection ceiling. Equivalently, it can be read as a discount on
later samples and so penalizes sample-greedy ensembles.

\paragraph{g-pass@$n_{\alpha}$.}
The threshold-pass variant: $\textrm{g-pass@}n_\alpha$ is the expected
indicator that at least a fraction $\alpha \in (0,1]$ of $n$ random
samples is correct. $\alpha = 1$ recovers $\textit{pass}^n$, and
$\alpha = 1/n$ recovers $\textrm{pass@}n$; we report the standard
$\alpha \in \{0.25,0.5,0.75,1.0\}$ ladder in the per-dataset reports of
\texttt{nl2smt.analyze}, but only the endpoints in the main tables below.

\subsection{Agent-Reliability Measures}
\label{app:metrics:agent}

These metrics compare the agent's behavior conditional on internal events.
As established in \S\ref{app:bg:agent}, $\kappa_x^{(r)}$ is the
\emph{commitment} bit, namely the JSONL \texttt{verified} field: it is
$1$ iff the agent terminated the loop by producing an answer rather than
hitting $T_{\max}$. $\chi_x^{(r)}$ is the correctness bit.
$\kappa_x^{(r)}=1,\chi_x^{(r)}=0$ is the \emph{overcommit} cell (the
agent declared confidence in a wrong answer) and is the operative
quantity in the soundness metric.

\paragraph{Soundness.}
\begin{equation}
\textrm{soundness}
=
\mathbb{P}(\chi = 1 \mid \kappa = 1)
=
\frac{\sum_{x,r} \chi_x^{(r)} \kappa_x^{(r)}}
     {\sum_{x,r} \kappa_x^{(r)}}.
\label{eq:soundness}
\end{equation}
Among attempts where the agent chose to finish instead of refining
further, what fraction were correct?
Soundness is a measure of \emph{decision quality}: it asks whether the
agent's internal stopping criterion is well-aligned with ground truth.
Low soundness indicates an agent that ``thinks it's done'' while still
wrong; a drop in soundness during post-training is diagnostic of
training-induced overconfidence, typically introduced by SFT on
demonstration traces that all terminate successfully.

\paragraph{Completeness.}
\begin{equation}
\textrm{completeness}
=
\mathbb{P}(\kappa = 1 \mid \chi = 1).
\label{eq:completeness}
\end{equation}
Among correct attempts, what fraction terminated voluntarily? Low
completeness is the dual failure to soundness: the agent silently arrives
at a correct answer but does not commit, instead burning iterations until
it hits $T_{\max}$. In practice every reported configuration in this work
has $\textrm{completeness} \geq 0.999$ because Qwen reliably emits a
\texttt{finish} terminator when satisfied. We retain completeness in the
table for diagnostic value and because failure modes on novel datasets
could plausibly violate it.

\paragraph{Recovery.}
\begin{equation}
\textrm{recovery}
=
\mathbb{P}\bigl(\chi = 1 \mid \text{first SMT call errored}\bigr).
\label{eq:recovery}
\end{equation}
The conditioning event is precisely
\[
\{\texttt{success}(z_x^{(r,1)}) = \textsc{False}\}
\]
or the first call's \texttt{sat result} was missing; i.e., the very
first Z3 invocation either failed to parse, ran out of resources, or did
not return a verdict. \emph{Recovery} measures the agent's robustness to
early failure: a high value indicates the agent diagnoses the error from
Z3's stderr text, rewrites the SMT, and eventually arrives at the right
answer; a low value indicates the agent commits to a wrong answer after
the first bad call. The difference
$\textrm{recovery} - \textrm{avg@}N$ measures the \emph{conditional}
penalty of a first-call failure, with the unconditional accuracy as
baseline. 

\paragraph{Expected Calibration Error.}
With only a binary commitment signal $\kappa$ and binary correctness
$\chi$, ECE collapses to the absolute gap between the per-attempt
commitment rate and the per-attempt accuracy:
\begin{equation}
\textrm{ECE}
=
\bigl|\,\overline{\kappa} - \overline{\chi}\,\bigr|.
\label{eq:ece}
\end{equation}
For a well-calibrated agent, the rate at which it commits matches the
rate at which it is right. Lower values are better; the metric becomes a
binned ECE if a continuous confidence signal is supplied via
\texttt{att["confidence"]}, but none of our checkpoints expose one.

\paragraph{Tool efficiency and turn efficiency.}
Let $S_x^{(r)}$ be the number of SMT calls and $T_x^{(r)}$ the number of
agent iterations in attempt $(x,r)$. Then
\begin{equation}
\begin{gathered}
\textrm{tool eff.}
=
\frac{\textrm{avg@}N}{\overline{S}},
\qquad
\textrm{turn eff.}
=
\frac{\textrm{avg@}N}{\overline{T}},
\\
\overline{S}
=
\frac{1}{MK}\sum_{x,r} S_x^{(r)},
\qquad
\overline{T}
=
\frac{1}{MK}\sum_{x,r} T_x^{(r)}.
\end{gathered}
\end{equation}
Both ratios are dimensionally \textsc{correctness}/\textsc{compute-unit}:
tool efficiency is accuracy per SMT call, the marginal value of
\emph{querying the solver}; turn efficiency is accuracy per agent
iteration, the marginal value of \emph{calling the model}. Because
$T_x^{(r)} \geq S_x^{(r)} + 1$, each SMT call consumes an iteration and
the final answer takes another, turn efficiency is always the smaller of
the two; the gap reflects how much non-solver agent work is being done
per problem.

\paragraph{Wasted turns.}
For each correct attempt, let $\iota_x^{(r)}$ denote the smallest index
$i$ such that the $i$-th SMT call in the trace has
\[
\rho_x^{(r,i)}
\in
\{\textsc{sat},\textsc{unsat}\},
\]
i.e., the earliest call that emitted a real Z3 verdict. Then
\begin{equation}
\begin{split}
\textrm{wasted turns}
&=
\mathbb{E}_{x,r:\chi_x^{(r)}=1}
\\
&\quad
\left[
\max\{0, S_x^{(r)}-\iota_x^{(r)}-1\}
\right],
\end{split}
\label{eq:wasted}
\end{equation}
the mean number of \emph{additional} SMT calls the agent issued
\emph{after} an already-correct call's verdict was returned. The metric
captures retroactive exploration: it is invariant to how hard a problem
was, only to whether the agent kept poking after the answer was implicitly
already known. Lower is better; a value of zero indicates the agent
commits the moment it has a real verdict.

\paragraph{Unique skeletons.}
For each problem we compute the SMT \emph{skeleton}, the sequence of
top-level command tokens of each call,
\[
(\texttt{declare-const}, \texttt{assert}, \texttt{check-sat}, \ldots),
\]
and count the number of distinct skeletons across the $K$ attempts. The
reported value is the mean of this count over $\mathcal{D}$; we also
report the mean pairwise Jaccard distance between the skeleton sets of
two attempts on the same problem. Both are diversity measures: large
values mean the agent explores qualitatively different solution shapes on
resampling; small values mean it produces structurally homogeneous
attempts. 

\subsection{SMT-Program Quality}
\label{app:metrics:smt}

Sample-based correctness metrics measure \emph{what answer} the agent
produced; SMT-program quality metrics measure \emph{how it got there}.
Each is a function of the SMT-LIB code and Z3 verdicts in the trace,
defined first per-call, then per-attempt via the lift in
Table~\ref{tab:row_quality_lift}, then aggregated over $\mathcal{D}$.
The per-attempt lift is materially important: an agent that emits a
strong SMT call \emph{somewhere} in its trace should not be penalized for
also issuing weak exploratory calls, so we use disjunctions, e.g.,
``any non-trivial,'' ``any grounded,'' and ``any real verdict,'' rather
than averages over calls.

\begin{table}[t]
\centering
\scriptsize
\begin{tabularx}{\linewidth}{@{}>{\ttfamily}l>{\raggedright\arraybackslash}X@{}}
\toprule
\normalfont\textbf{Attribute} & \textbf{Definition} \\
\midrule
any non trivial
&
$\bigvee_i \neg\,\texttt{is trivial}(z_x^{(r,i)})$
\\
any grounded
&
$\bigvee_i
\left(
\texttt{grounding}(z_x^{(r,i)},x) \geq \lambda_{\mathrm{gnd}}
\right)$
\\
any real verdict
&
$\bigvee_i
\left(
\texttt{sat result}_i
\in
\{\textsc{sat},\textsc{unsat},\textsc{unknown}\}
\right)$
\\
any unsat
&
$\bigvee_i
\left(
\texttt{sat result}_i = \textsc{unsat}
\right)$
\\
first pass valid
&
$\texttt{sat result}_1
\in
\{\textsc{sat},\textsc{unsat},\textsc{unknown}\}$
\\
max grounding
&
$\max_i \texttt{grounding}(z_x^{(r,i)},x)$
\\
\bottomrule
\end{tabularx}
\caption{Per-attempt structural attributes computed from the sequence of
SMT calls
$(z_x^{(r,1)},\ldots,z_x^{(r,S_x^{(r)})})$
in trace $\tau_x^{(r)}$. The grounding threshold is
$\lambda_{\mathrm{gnd}}=0.25$. \texttt{is trivial} is the predicate
defined in \S\ref{app:metrics:smt:trivial}.}
\label{tab:row_quality_lift}
\end{table}

\subsubsection{Grounding}
\label{app:metrics:smt:grounding}

The grounding score quantifies how much of the problem's actual content the
agent transferred into the formal SMT representation. We extract
\emph{problem entities} $E(x)$ from the problem text $x$ via the regular
expression \texttt{[A-Za-z][A-Za-z0-9 ]\{1,\}} after lower-casing, then
remove a small fixed list of $\sim 50$ English stopwords plus generic task
scaffolding tokens, e.g., \texttt{question}, \texttt{answer},
\texttt{true}, \texttt{false}, and \texttt{option}. For an SMT call $z$,
the grounding score is the fraction of problem entities that appear as
substrings in the lower-cased SMT body:
\begin{equation}
\begin{split}
\mathrm{gnd}(z,x)
&=
\frac{1}{|E(x)|}
\sum_{e \in E(x)}
\mathbb{1}\!\left[
e \in \operatorname{lc}(z)
\right],
\end{split}
\label{eq:grounding}
\end{equation}
where $\operatorname{lc}(z)$ is the lower-cased SMT body. We use the
convention $\mathrm{gnd}=0$ when $|E(x)|=0$, which is
vanishingly rare in our datasets. Because the predicate ``$e$ is a
substring of $z$'' is permissive (it counts both
\texttt{declare-const e Int} and any occurrence of $e$ in a comment or
sort name), grounding is a \emph{liberal} measure of semantic transfer
rather than a strict one, intentionally biased toward false positives so
that the metric does not reward \emph{intentional} obfuscation. The
reported \texttt{mean grounding} is the per-attempt
\texttt{max grounding} averaged over $\mathcal{D}$.

A grounding score of $0.05$ on a problem with twenty named entities
indicates the SMT references one of them: this is the
\emph{decorative-SMT} regime in which the agent has constructed a
formally valid program that does not actually express the problem. A
score of $0.5$ indicates the SMT references roughly half the entities,
sufficient for the program to be non-vacuous. The threshold
$\lambda_{\mathrm{gnd}}=0.25$ used in
Table~\ref{tab:row_quality_lift} is empirically set: it cleanly
separates inspection-verified decorative programs from inspection-verified
content-loaded ones on a held-out development sample.

\subsubsection{Triviality}
\label{app:metrics:smt:trivial}

An SMT call $z$ is \emph{trivial} if any of the following hold:
\begin{enumerate}
  \setlength{\itemsep}{0.1em}
  \item $z$ is empty or whitespace.
  \item $z$ contains no \texttt{declare-const}, \texttt{declare-fun},
  or \texttt{define-fun}, and contains fewer than two
  \texttt{(assert ...)} forms.
  \item Every \texttt{(assert ...)} form in $z$ matches a literal
  tautology or contradiction:
  $(\texttt{assert true})$, $(\texttt{assert (= n n)})$ with $n$ a
  numeric literal, $(\texttt{assert (not true)})$, or
  $(\texttt{assert (not false)})$.
  \item $z$ has fewer than $40$ non-whitespace characters and at most
  one assertion.
\end{enumerate}
The first three cases catch \emph{semantically empty} SMT, i.e., programs that
contain Z3 syntax but no problem content. The fourth case catches
\emph{stub} programs that initiate an SMT-LIB block but abandon it before
encoding anything substantive. The \texttt{trivial smt rate} reported
in tables is the fraction of attempts whose \emph{entire} SMT call
sequence is trivial:
\begin{equation}
\begin{split}
\mathrm{triv}_{\mathrm{SMT}}
&=
\frac{1}{N_{\mathrm{SMT}}}
\sum_{(x,r)}
\mathbb{1}\!\left[
\bigwedge_i \operatorname{triv}(z_x^{(r,i)})
\right],
\end{split}
\end{equation}
where $N_{\mathrm{SMT}}$ is the number of attempts with at least one SMT
call. A non-zero trivial-SMT rate is diagnostic of \emph{performative
SMT}: the agent emits Z3 calls to satisfy the tool-calling prior of its
training distribution without actually using the solver to reason. 

\subsubsection{First-Pass Z3 Validity}
\label{app:metrics:smt:first_pass}

The fraction of attempts, over those with at least one SMT call, whose
very first invocation produced a real Z3 verdict:
\begin{equation}
\begin{gathered}
\mathcal{V}_{\mathrm{Z3}}
=
\{\textsc{sat},\textsc{unsat},\textsc{unknown}\},
\\
\mathcal{A}_{\mathrm{rv}}
=
\{(x,r): \rho_x^{(r,1)} \in \mathcal{V}_{\mathrm{Z3}}\},
\\
\mathcal{A}_{\mathrm{SMT}}
=
\{(x,r): S_x^{(r)} \geq 1\},
\\
\mathrm{fpv}_{\mathrm{Z3}}
=
\frac{|\mathcal{A}_{\mathrm{rv}}|}
     {|\mathcal{A}_{\mathrm{SMT}}|}.
\end{gathered}
\end{equation}
This isolates whether the agent's very first SMT-LIB emission is
syntactically well-formed under SMT-LIB v2 and well-typed under the
declared theory. A delta on this metric during post-training reflects
improvements in surface-level SMT syntax (fewer malformed parentheses,
undeclared symbols, and theory mismatches), not in reasoning depth.

\subsubsection{Verified-Correct Rate and its Unsat Restriction}
\label{app:metrics:smt:vcr}

We now define the strictest accuracy measure used in this work. An attempt
is \emph{verified-correct} if the conjunction
\begin{equation}
\begin{aligned}
\mathrm{VC}_x^{(r)}
:={}&
\chi_x^{(r)}
\\
&\wedge
\texttt{any real verdict}_x^{(r)}
\\
&\wedge
\texttt{any grounded}_x^{(r)}
\\
&\wedge
\texttt{any non trivial}_x^{(r)}
\end{aligned}
\label{eq:vc_attempt}
\end{equation}
holds, and verified-correct under \textsc{unsat} if it further satisfies
$\texttt{any unsat}_x^{(r)}$. The aggregate rates are
\begin{equation}
\begin{aligned}
\mathrm{vcr}
&=
\frac{1}{MK}\sum_{x,r} \mathrm{VC}_x^{(r)},
\\
\mathrm{vcr}_{\textsc{unsat}}
&=
\frac{1}{MK}\sum_{x,r}
\left(
\mathrm{VC}_x^{(r)}
\wedge
\texttt{any unsat}_x^{(r)}
\right).
\end{aligned}
\end{equation}
$\mathrm{vcr}$ filters out four distinct failure modes that can all
produce a passing $\chi_x^{(r)}=1$:
\begin{itemize}
  \setlength{\itemsep}{0.1em}
  \item \emph{No real verdict}: the agent answered correctly without ever
  obtaining a \textsc{sat}/\textsc{unsat}/\textsc{unknown} verdict from
  Z3, i.e., all calls errored or produced no verdict.
  \item \emph{Decorative SMT}: the agent obtained a verdict but the
  program had grounding below $\lambda_{\mathrm{gnd}}$, i.e., the SMT did
  not reference the problem.
  \item \emph{Trivial SMT}: every call was structurally trivial.
  \item \emph{The conjunction held but on a vacuous verdict}: e.g., a
  \textsc{sat} verdict on the empty constraint set.
\end{itemize}
Of these the most dangerous is the no-real-verdict case: it indicates a
model that answers correctly via natural-language reasoning while only
\emph{performing} SMT use. The post-training $\mathrm{vcr}$ delta
measures the shift from performative to real solver use.

$\mathrm{vcr}_{\textsc{unsat}}$ goes further by requiring that at least
one SMT call in the trace returned \textsc{unsat}. By the discussion in
\S\ref{app:bg:z3}, this is the \emph{refutational-proof} restriction:
the agent must have attempted to derive a contradiction from the negation
of its candidate answer, and the solver must have confirmed it. The pair
$(\mathrm{vcr},\mathrm{vcr}_{\textsc{unsat}})$ therefore separates
correctness into three regimes: (i) correct but performative; (ii)
verified-correct via \textsc{sat} witness, i.e., model-finding; and
(iii) verified-correct via \textsc{unsat}, i.e., refutational proof. 

\subsubsection{Z3 Outcome Distribution}
\label{app:metrics:smt:outcomes}

For diagnostic and downstream analysis purposes, we additionally report the
distribution of Z3 outcomes across every SMT call in the sweep:
\[
\begin{gathered}
\omega^{\mathrm{Z3}}
=
\bigl(
\omega_{\textsc{sat}},
\omega_{\textsc{unsat}},
\omega_{\textsc{unknown}},
\\
\omega_{\textsc{error}},
\omega_{\textsc{no-verdict}}
\bigr),
\end{gathered}
\]
with $\omega^{\mathrm{Z3}}_u$ the fraction of all SMT calls whose
\texttt{sat result} took value $u$. We treat \textsc{no verdict}, where
the call succeeded but no \texttt{sat result} token appeared in stdout,
as distinct from \textsc{error}, where the call raised an exception or
returned a non-zero exit.

\section{Comparison to Open-Weight Baselines}
\label{app:open-weight-comparisons}

We additionally compare the DPO-tuned Sequent-Prover checkpoints against
external open-weight baselines evaluated with the same SMT-scratchpad harness.
The external baselines are Gemma-4B, GPT-OSS-20B, and GPT-OSS-120B. All deltas are Qwen-DPO minus the external model.  Positive values indicate that the Qwen-DPO checkpoint is higher,
except for lower-is-better diagnostic rates, such as the trivial-SMT rate, where the
raw signed delta is still shown.

\paragraph{Aggregate correctness.}
Table~\ref{tab:open-weight-correctness} reports macro-averaged deltas over the
shared canonical datasets.  The 4B comparison shows a large separation from
Gemma-4B: Qwen-4B-DPO improves avg@$N$ by $+0.647$, pass@4 by $+0.524$, and
\passhat{4} by $+0.640$.  The GPT-OSS-20B comparisons show the same pattern at a
larger external baseline: Qwen-27B-DPO improves avg@$N$ by $+0.326$ and
\passhat{4} by $+0.648$, while Qwen-9B-DPO improves avg@$N$ by $+0.287$ and
\passhat{4} by $+0.567$.  Against GPT-OSS-120B, the comparison is closer and
therefore more informative: Qwen-27B-DPO remains positive on all correctness
aggregates, including $+0.027$ avg@$N$, $+0.021$ pass@4, and $+0.036$ \passhat{4};
Qwen-9B-DPO is close on coverage, with $+0.006$ pass@4 and $+0.002$ AUC@4,
while GPT-OSS-120B has the advantage on avg@$N$ and \passhat{4}.

\begin{table*}[t]
\centering
\small
\setlength{\tabcolsep}{4.6pt}
\begin{tabular}{@{}lrrrrrrr@{}}
\toprule
\textbf{Comparison} & \textbf{Shared} & \textbf{avg@$N$} & \textbf{pass@4} &
\textbf{\passhat{4}} & \textbf{maj@4} & \textbf{AUC@4} & \textbf{Cons.} \\
\midrule
Qwen-4B-DPO $-$ Gemma-4B       & 33 & +0.647 & +0.524 & +0.640 & +0.720 & +0.591 & +0.568 \\
Qwen-27B-DPO $-$ GPT-OSS-20B   & 35 & +0.326 & +0.067 & +0.648 & +0.408 & +0.151 & +0.699 \\
Qwen-9B-DPO $-$ GPT-OSS-20B    & 33 & +0.287 & +0.053 & +0.567 & +0.372 & +0.130 & +0.625 \\
\midrule
Qwen-27B-DPO $-$ GPT-OSS-120B  & 35 & +0.027 & +0.021 & +0.036 & +0.028 & +0.023 & +0.017 \\
Qwen-9B-DPO $-$ GPT-OSS-120B   & 33 & \loss{$-$0.010} & +0.006 & \loss{$-$0.039} & \loss{$-$0.008} & +0.002 & \loss{$-$0.054} \\
\bottomrule
\end{tabular}
\caption{Macro-averaged correctness deltas against external open-weight
baselines.  Deltas are computed over shared canonical datasets and are reported
as Qwen-DPO minus the external model.  Cons. is \passhat{4}/pass@4. \loss{Red} marks a deficit.}
\label{tab:open-weight-correctness}
\end{table*}

\paragraph{Verifier-use diagnostics.}
Table~\ref{tab:open-weight-verifier} compares solver-facing behavior.  The
largest gains are in the verified-correct rate, UNSAT-restricted verified-correct
rate, and grounding.  Qwen-4B-DPO improves verified-correct rate by $+0.615$
relative to Gemma-4B.  Against GPT-OSS-20B, both Qwen-27B-DPO and Qwen-9B-DPO
improve the verified-correct rate by $+0.487$.  The GPT-OSS-120B comparison is
particularly useful because it separates answer-level strength from
verifier-use quality: Qwen-27B-DPO improves verified-correct rate by $+0.341$
and grounding by $+0.274$, while Qwen-9B-DPO improves verified-correct rate by
$+0.349$ and grounding by $+0.299$.  Trivial-SMT rate decreases in all five
comparisons, indicating that the DPO checkpoints emit fewer structurally empty
or decorative SMT programs.

\begin{table*}[t]
\centering
\small
\setlength{\tabcolsep}{4.4pt}
\begin{tabular}{@{}lrrrrrrr@{}}
\toprule
\textbf{Comparison} & \textbf{Sound.} & \textbf{Recov.} & \textbf{VCR} &
\textbf{VCR-U} & \textbf{FP-Z3} & \textbf{Triv. SMT $\downarrow$} & \textbf{Ground.} \\
\midrule
Qwen-4B-DPO $-$ Gemma-4B       & +0.444 & +0.490 & +0.615 & +0.399 & +0.164 & \gain{$-$0.016} & +0.399 \\
Qwen-27B-DPO $-$ GPT-OSS-20B   & +0.206 & +0.181 & +0.487 & +0.345 & +0.034 & \gain{$-$0.032} & +0.329 \\
Qwen-9B-DPO $-$ GPT-OSS-20B    & +0.172 & +0.152 & +0.487 & +0.321 & +0.023 & \gain{$-$0.031} & +0.353 \\
\midrule
Qwen-27B-DPO $-$ GPT-OSS-120B  & \loss{$-$0.010} & \loss{$-$0.020} & +0.341 & +0.196 & +0.004 & \gain{$-$0.011} & +0.274 \\
Qwen-9B-DPO $-$ GPT-OSS-120B   & \loss{$-$0.036} & \loss{$-$0.032} & +0.349 & +0.179 & \loss{$-$0.008} & \gain{$-$0.011} & +0.299 \\
\bottomrule
\end{tabular}
\caption{Macro-averaged verifier-use deltas.  Sound. is
$P(\mathrm{correct}\mid\mathrm{finish})$; Recov. is correctness after an
initial SMT error; VCR is verified-correct rate; VCR-U additionally requires an
UNSAT verdict; FP-Z3 is first-pass Z3 validity; Triv. SMT is trivial-SMT rate,
where lower is better; Ground. is mean SMT grounding. Negative entries are \gain{green} where lower is better and \loss{red} otherwise.}
\label{tab:open-weight-verifier}
\end{table*}

\paragraph{Per-dataset breadth.}
Table~\ref{tab:open-weight-wlt} reports polarity-adjusted win/loss/tie counts
across shared datasets.  The Qwen-DPO checkpoints dominate the 4B and 20B
external baselines across the main sample-based metrics: \passhat{4} and maj@4
win on every shared dataset in the Gemma-4B and GPT-OSS-20B comparisons.  The
GPT-OSS-120B comparison is more balanced.  Qwen-27B-DPO wins avg@$N$ on 28 of
35 datasets and \passhat{4} on 24 of 35 datasets, with five \passhat{4} ties.
Qwen-9B-DPO remains competitive on coverage, winning pass@4 on 22 of 33
datasets, while GPT-OSS-120B has the stronger all-sample consistency profile.

\begin{table*}[t]
\centering
\scriptsize
\setlength{\tabcolsep}{3.6pt}
\begin{tabular}{@{}lrrrrrrr@{}}
\toprule
\textbf{Comparison} & \textbf{avg@$N$} & \textbf{pass@4} & \textbf{\passhat{4}} &
\textbf{maj@4} & \textbf{FP-Z3} & \textbf{Sound.} & \textbf{Recov.} \\
\midrule
Qwen-4B-DPO $-$ Gemma-4B       & 32/0/1 & 32/1/0 & 33/0/0 & 33/0/0 & 33/0/0 & 27/1/5 & 27/1/5 \\
Qwen-27B-DPO $-$ GPT-OSS-20B   & 35/0/0 & 33/1/1 & 35/0/0 & 35/0/0 & 35/0/0 & 29/1/5 & 27/2/6 \\
Qwen-9B-DPO $-$ GPT-OSS-20B    & 33/0/0 & 31/1/1 & 33/0/0 & 33/0/0 & 29/3/1 & 26/2/5 & 23/5/5 \\
\midrule
Qwen-27B-DPO $-$ GPT-OSS-120B  & 28/7/0 & 23/9/3 & 24/6/5 & 25/9/1 & 22/9/4 & 16/14/5 & 14/15/6 \\
Qwen-9B-DPO $-$ GPT-OSS-120B   & 16/17/0 & 22/8/3 & 6/26/1 & 15/16/2 & 3/29/1 & 8/20/5 & 6/21/6 \\
\bottomrule
\end{tabular}
\caption{Per-dataset win/loss/tie counts.  Each entry is W/L/T after metric
polarity is applied.  Higher is better for all displayed metrics.}
\label{tab:open-weight-wlt}
\end{table*}

\paragraph{Category-level trends.}
Table~\ref{tab:open-weight-categories} groups the avg@$N$ deltas by benchmark
category.  Qwen-DPO is substantially ahead of Gemma-4B and GPT-OSS-20B in every
category.  Against GPT-OSS-120B, Qwen-27B-DPO is strongest on AGIEval, NLI/MCQ,
and hard-reasoning categories, and remains close on arithmetic and free-form
QA.  The 9B-DPO comparison to GPT-OSS-120B is near parity for AGIEval,
arithmetic, hard reasoning, and NLI/MCQ, with larger gaps on spatial and
free-form QA tasks.

\begin{table*}[t]
\centering
\scriptsize
\setlength{\tabcolsep}{3.0pt}
\begin{tabular}{@{}lrrrrr@{}}
\toprule
\textbf{Category} & \textbf{4B$-$Gemma} & \textbf{27B$-$OSS20} &
\textbf{9B$-$OSS20} & \textbf{27B$-$OSS120} &
\textbf{9B$-$OSS120} \\
\midrule
AGIEval       & +0.795 & +0.376 & +0.331 & +0.069 & +0.024 \\
Arithmetic    & +0.686 & +0.304 & +0.283 & +0.021 & +0.001 \\
Free-form QA  & +0.611 & +0.363 & +0.312 & +0.002 & \loss{$-$0.049} \\
Hard reasoning& +0.659 & +0.346 & +0.317 & +0.035 & \loss{$-$0.003} \\
NLI/MCQ       & +0.647 & +0.346 & +0.301 & +0.044 & +0.003 \\
Other         & +0.816 & +0.307 & +0.285 & \loss{$-$0.008} & \loss{$-$0.022} \\
Spatial       & +0.341 & +0.143 & +0.110 & \loss{$-$0.037} & \loss{$-$0.071} \\
\bottomrule
\end{tabular}
\caption{Mean avg@$N$ delta by dataset category.  Values are macro-averages over
shared canonical datasets in each category.}
\label{tab:open-weight-categories}
\end{table*}

These external comparisons support two conclusions.  First,
Sequent-Prover DPO checkpoints are substantially stronger than the same-scale or
moderately larger open-weight baselines under the SMT-scratchpad evaluation,
with especially large advantages in \passhat{4} and verifier-grounded correctness.
Second, the comparison with GPT-OSS-120B shows that the 27B DPO checkpoint is
competitive with a much larger open-weight model on answer-level metrics, while
retaining a clear advantage in the use of executable, grounded solvers.

\section{Additional Analyses}
\label{app:additional}

This section gives full tables and details behind the main results: pool
composition and per-category results, the test-time sampling study, grounding
without comments, structural uncertainty of the generated SMT, teacher
support, and soundness.

\subsection{Pool Composition and Per-Category Results}
\label{app:add:categories}

Each benchmark contributes 200 problems, except ProofWriter, which contributes
54 because of a data-loader bug. This gives $36\times200+54=7{,}254$ problems
per family. Table~\ref{tab:pool-composition} groups the pool by category.
Table~\ref{tab:category-results} reports Base, \sft{}, and \dpo{} results within
each category. Values are micro-averages over the problems in each category,
using the same four trajectories per problem as the main results.

\begin{table}[h]
\centering
\small
\begin{tabular}{@{}lrrr@{}}
\toprule
\textbf{Category} & \textbf{Bench.} & \textbf{Problems} & \textbf{\%} \\
\midrule
Hard reasoning & 6  & 1,054 & 14.53 \\
Arithmetic     & 6  & 1,200 & 16.54 \\
Spatial        & 3  & 600   & 8.27 \\
NLI/MCQ        & 12 & 2,400 & 33.09 \\
Free-form QA   & 4  & 800   & 11.03 \\
Other          & 2  & 400   & 5.51 \\
AGIEval        & 4  & 800   & 11.03 \\
\midrule
Total          & 37 & 7,254 & 100.00 \\
\bottomrule
\end{tabular}
\caption{Composition of the 37-benchmark evaluation pool.}
\label{tab:pool-composition}
\end{table}

\begin{table*}[t]
\centering
\small
\setlength{\tabcolsep}{4.5pt}
\begin{tabular}{@{}llrrrrrr@{}}
\toprule
& & \multicolumn{3}{c}{\textbf{avg@$N$ $\uparrow$}} & \multicolumn{3}{c}{\textbf{\passhat{4} $\uparrow$}} \\
\cmidrule(lr){3-5}\cmidrule(l){6-8}
\textbf{Family} & \textbf{Category} & \base & \sft & \dpo & \base & \sft & \dpo \\
\midrule
4B & Hard reasoning & 0.630 & 0.920 & 0.931 & 0.346 & 0.820 & 0.839 \\
   & Arithmetic     & 0.899 & 0.984 & 0.987 & 0.691 & 0.952 & 0.964 \\
   & Spatial        & 0.735 & 0.867 & 0.895 & 0.372 & 0.670 & 0.708 \\
   & NLI/MCQ        & 0.899 & 0.977 & 0.985 & 0.688 & 0.926 & 0.953 \\
   & Free-form QA   & 0.956 & 0.988 & 0.992 & 0.841 & 0.960 & 0.971 \\
   & Other          & 0.950 & 0.988 & 0.997 & 0.815 & 0.950 & 0.988 \\
   & AGIEval        & 0.900 & 0.955 & 0.970 & 0.729 & 0.863 & 0.911 \\
\midrule
9B & Hard reasoning & 0.853 & 0.938 & 0.951 & 0.687 & 0.859 & 0.871 \\
   & Arithmetic     & 0.979 & 0.988 & 0.989 & 0.945 & 0.970 & 0.973 \\
   & Spatial        & 0.896 & 0.891 & 0.916 & 0.707 & 0.712 & 0.777 \\
   & NLI/MCQ        & 0.968 & 0.987 & 0.987 & 0.887 & 0.959 & 0.960 \\
   & Free-form QA   & 0.987 & 0.993 & 0.997 & 0.949 & 0.978 & 0.989 \\
   & Other          & 0.973 & 0.992 & 0.996 & 0.905 & 0.968 & 0.988 \\
   & AGIEval        & 0.953 & 0.976 & 0.976 & 0.863 & 0.911 & 0.914 \\
\midrule
27B & Hard reasoning & 0.967 & 0.974 & 0.992 & 0.924 & 0.926 & 0.976 \\
    & Arithmetic     & 0.996 & 0.996 & 0.998 & 0.993 & 0.988 & 0.994 \\
    & Spatial        & 0.962 & 0.945 & 0.974 & 0.912 & 0.858 & 0.923 \\
    & NLI/MCQ        & 0.998 & 0.998 & 0.999 & 0.993 & 0.994 & 0.995 \\
    & Free-form QA   & 1.000 & 0.999 & 0.999 & 1.000 & 0.996 & 0.995 \\
    & Other          & 0.999 & 0.999 & 1.000 & 0.998 & 0.998 & 1.000 \\
    & AGIEval        & 0.998 & 0.993 & 0.997 & 0.988 & 0.980 & 0.990 \\
\bottomrule
\end{tabular}
\caption{Per-category results for Base, \sft{}, and \dpo{} checkpoints. Values are micro-averages over the problems in each category.}
\label{tab:category-results}
\end{table*}

Hard reasoning has the largest within-category gain at 4B (+0.493
\passhat{4}). NLI/MCQ, the largest category, also
contributes strongly. At 27B, most categories are near ceiling. The remaining
gains are in hard reasoning and spatial reasoning, with small changes on
saturated categories.

\subsection{Post-Training versus Test-Time Sampling}
\label{app:add:ttc}

We compare post-training with a larger test-time sampling budget. We fix 180
held-out problems, 30 each from ProntoQA, FOLIO, AR-LSAT, GSM8K, StepGame, and
ARC-Challenge. We draw 16 trajectories per problem from the 4B base model and
from the post-trained 4B model. We also draw four trajectories per problem with
decomposed prompting, least-to-most prompting, and self-refinement on the base
model. All methods share the solver harness, temperature 0.9, a limit of ten
agent iterations, the answer checker, and the problem list. The study contains
7,920 trajectories. Table~\ref{tab:ttc} reports equal-weight macro-averages over
the six families. Intervals are stratified 95\% bootstrap intervals over
problems (10,000 resamples).

\begin{table*}[t]
\centering
\small
\setlength{\tabcolsep}{4.5pt}
\begin{tabular}{@{}lrlrrrrr@{}}
\toprule
\textbf{Method} & \textbf{$K$} & \textbf{Accuracy [95\% CI] $\uparrow$} & \textbf{pass@$K$ $\uparrow$} &
\textbf{\passhat{K} $\uparrow$} & \textbf{Stab. $\uparrow$} & \textbf{Tokens/problem $\downarrow$} & \textbf{Cost $\downarrow$} \\
\midrule
Base + self-consistency & 1  & 0.657 [0.623, 0.689] & 0.657 & 0.657 & 1.000 & 27,729  & 1.7$\times$ \\
Base + self-consistency & 2  & 0.657 [0.623, 0.689] & 0.780 & 0.533 & 0.683 & 55,458  & 3.3$\times$ \\
Base + self-consistency & 4  & 0.657 [0.623, 0.689] & 0.848 & 0.395 & 0.466 & 110,915 & 6.6$\times$ \\
Base + self-consistency & 8  & 0.657 [0.622, 0.689] & 0.887 & 0.257 & 0.290 & 221,831 & 13.3$\times$ \\
Base + self-consistency & 16 & 0.657 [0.623, 0.689] & 0.911 & 0.139 & 0.152 & 443,662 & 26.5$\times$ \\
\midrule
Base + decomposed       & 4  & 0.590 [0.550, 0.629] & 0.828 & 0.267 & 0.322 & 117,070 & 7.0$\times$ \\
Base + least-to-most    & 4  & 0.621 [0.578, 0.661] & 0.828 & 0.317 & 0.383 & 121,933 & 7.3$\times$ \\
Base + self-refine      & 4  & 0.772 [0.733, 0.811] & 0.928 & 0.572 & 0.617 & 219,491 & 13.1$\times$ \\
\midrule
\ours Post-trained 4B         & 1  & 0.869 [0.836, 0.899] & 0.869 & 0.869 & 1.000 & 16,718  & 1.0$\times$ \\
\ours Post-trained 4B         & 2  & 0.869 [0.835, 0.899] & 0.924 & 0.813 & 0.880 & 33,437  & 2.0$\times$ \\
\ours Post-trained 4B         & 4  & 0.869 [0.836, 0.900] & 0.948 & 0.743 & 0.784 & 66,873  & 4.0$\times$ \\
\ours Post-trained 4B         & 8  & 0.869 [0.836, 0.900] & 0.960 & 0.663 & 0.690 & 133,747 & 8.0$\times$ \\
\ours Post-trained 4B         & 16 & 0.869 [0.835, 0.900] & 0.967 & 0.572 & 0.592 & 267,493 & 16.0$\times$ \\
\bottomrule
\end{tabular}
\caption{Post-training versus test-time sampling on 180 held-out problems from six benchmark families. Cost is token use relative to the post-trained model at $K=1$.}
\label{tab:ttc}
\end{table*}

Beyond the results in Section~\ref{sec:results:ttc}, at matched $K=4$ the
post-trained model also improves accuracy by 21.2 points and majority-vote
accuracy by 9.9 points over base self-consistency. Against the base model at
$K=16$, majority-vote accuracies are statistically similar, but the base model
needs 6.6$\times$ more tokens to get there. Decomposed and least-to-most
prompting do not improve over base self-consistency on this mixture.

\subsection{Grounding without Comments}
\label{app:add:grounding}

Grounding (Eq.~\ref{eq:grounding}) counts entity names anywhere in the SMT
body, including comments. Comments often hold the only readable mapping from
short SMT symbols to problem entities. For example, an AR-LSAT trace may name
people \texttt{J} and \texttt{K} in code and explain them in comments. Still, a
model could in principle raise grounding by writing entity names in comments.
Grounding is not part of any training reward; it is only a filter for Gold
traces. To test this concern, we strip comments and recompute grounding on about
50k trajectories from the runs in the main paper.
Table~\ref{tab:grounding-nocomment} shows that post-trained models still have
higher grounding than their base models.

\begin{table}[h]
\centering
\small
\begin{tabular}{@{}lrr@{}}
\toprule
\textbf{Family} & \textbf{\sft{} $-$ Base} & \textbf{\dpo{} $-$ Base} \\
\midrule
4B  & \gain{+0.074} & \gain{+0.081} \\
9B  & \gain{+0.051} & \gain{+0.051} \\
27B & \gain{+0.039} & \gain{+0.036} \\
\bottomrule
\end{tabular}
\caption{Change in mean grounding after removing SMT comments.}
\label{tab:grounding-nocomment}
\end{table}

\subsection{Structural Uncertainty of SMT Programs}
\label{app:add:pcfg}

Full semantic validation of a formalization is not possible in general (see
Limitations). As a symbolic proxy, we follow \citet{ganguly2026grammars}, who
show that structural variation in sampled SMT-LIB programs predicts
autoformalization errors. We ask whether post-training reduces this structural
uncertainty.

For task $x$ and independent trajectory $j$, let $s_{x,j}$ be its final
parseable SMT call. We parse $s_{x,j}$ with the SMT-LIB ANTLR grammar and count
each grammar production $r$, giving counts $c_{x,j}(r)$ and the distribution
$P_{x,j}(r)=c_{x,j}(r)/\sum_{r'}c_{x,j}(r')$. The lexer drops comments, and all
identifiers map to one token class. The measure is unchanged by
comments or renaming, but it is sensitive to declarations, quantifiers,
operators, and term structure. We compare two distributions with the
Jensen--Shannon divergence $\mathrm{JSD}$ in base 2, so
$\mathrm{JSD}\in[0,1]$. A value of zero means equal production frequencies. It
does not imply identical syntax trees or semantics.

\paragraph{Across trajectories.} For trajectory $j$, we pool the counts of its
peers into $\bar{P}_{x,-j}(r)=\sum_{i\neq j}c_{x,i}(r)/\sum_{i\neq j}\sum_{r'}c_{x,i}(r')$
and score $D_{x,j}=\mathrm{JSD}(P_{x,j}\,\|\,\bar{P}_{x,-j})$. The scored program
never enters its own reference. We require at least two parsed peers and
average $D_{x,j}$ within each task to get $D_x$.

\paragraph{Within a trajectory.} For a trajectory with $L\geq2$ parseable SMT
calls with distributions $P^{(1)},\ldots,P^{(L)}$, we measure first--last drift
$\mathrm{JSD}(P^{(1)}\,\|\,P^{(L)})$ and mean adjacent drift
$\frac{1}{L-1}\sum_{\ell=1}^{L-1}\mathrm{JSD}(P^{(\ell)}\,\|\,P^{(\ell+1)})$.

For each measure we average over eligible trajectories in a task and report
the paired task-level change $\Delta=\mathbb{E}_x[D_x^{\mathrm{tuned}}-D_x^{\mathrm{base}}]$.
Negative values mean lower structural variation after post-training.

\begin{table}[h]
\centering
\small
\setlength{\tabcolsep}{3pt}
\resizebox{\linewidth}{!}{%
\begin{tabular}{@{}lrrr@{}}
\toprule
& \textbf{Across traj.} & \multicolumn{2}{c}{\textbf{Within traj.}} \\
\cmidrule(lr){2-2}\cmidrule(l){3-4}
\textbf{Comparison} & \textbf{$\Delta$ [95\% CI] $\downarrow$} & \textbf{First--last $\downarrow$} & \textbf{Adjacent $\downarrow$} \\
\midrule
4B Base$\to$\sft  & $-0.023$ [$-0.025$, $-0.021$] & $-0.029$ & $-0.011$ \\
4B Base$\to$\dpo  & $-0.019$ [$-0.021$, $-0.016$] & $-0.027$ & $-0.008$ \\
9B Base$\to$\sft  & $-0.019$ [$-0.021$, $-0.017$] & $-0.024$ & $-0.010$ \\
9B Base$\to$\dpo  & $-0.016$ [$-0.018$, $-0.015$] & $-0.026$ & $-0.010$ \\
27B Base$\to$\sft & $-0.008$ [$-0.010$, $-0.006$] & $-0.006$ & $-0.005$ \\
27B Base$\to$\dpo & $-0.004$ [$-0.006$, $-0.002$] & $-0.011$ & $-0.006$ \\
\bottomrule
\end{tabular}}
\caption{Paired change in structural (PCFG production) divergence over 1,382 tasks per comparison. Across-trajectory intervals use 4,000 bootstrap resamples, stratified by dataset. Within-trajectory columns are point estimates.}
\label{tab:pcfg}
\end{table}

In Table~\ref{tab:pcfg}, the drop in peer divergence is largest at 4B and 9B
and smaller at 27B, whose base model is already stable. Both within-trajectory
drifts also fall in all six comparisons, so tuned models make smaller
structural revisions during repair.

\subsection{Teacher Support and Student Capability}
\label{app:add:teacher}

SFT imitates the teacher conditioned on success,
$p_T(\tau\mid x,A(x,y^\star,\tau)=1)$ (Section~\ref{sec:method}). With $K_T$
samples, at least one accepted trace appears with probability
$1-(1-Z_T(x))^{K_T}$. The student is bounded by the teacher's
\emph{support} over successful strategies.
The same idea underlies expert iteration, where an expensive search procedure
produces behavior that a policy learns to reproduce~\citep{ren2025deepseek}. It
explains why the students exceed the teacher's all-sample reliability on AR-LSAT
(Section~\ref{sec:results:benchmarks}, Table~\ref{tab:teacher-acceptance}).
The students learn reusable steps, such as datatype encodings and push/pop
option testing (Appendix~\ref{sec:datatypes},~\ref{sec:pushpop}). The teacher is weak on some SMT fragments,
such as fixed-size bit-vector arithmetic. Our benchmarks do not test these
fragments, and a stronger teacher there would extend the same recipe.

\subsection{Soundness after SFT}
\label{app:add:soundness}

Soundness is the accuracy of attempts on which the agent chooses to finish
(Eq.~\ref{eq:soundness}). All SFT demonstrations are successful traces that end
cleanly. This may teach the model to commit early, which would lower soundness.
We do not see a broad drop in reliability that would confirm this effect.
After DPO, soundness improves, together with fewer malformed SMT calls and more
refutational solver use (Table~\ref{tab:smt-quality}).

\section{Qualitative Improvements}

The quantitative improvements we observe in our metrics, such as verified correct rate with unsat (i.e., appropriate proof by contradiction) correlate to formalizations that, on manual inspection, follow the problem statement much more closely. Throughout the harness and metric development, as well as during intermediate checkpoints, we took care to ensure there is no reward hacking and quantitative improvements actually manifest via qualitative improvements in formal artifact quality. In this section, we attempt to elucidate a few learnt capabilities in our trained models.

\subsection{Capability 1: SMT-LIB language fluency}
\label{sec:fluency}

The most basic gap is that the 4B Qwen Base model emits malformed SMT-LIB on the
first attempt and rarely recovers, whereas Sequent-Prover-4B recovers in one
turn. We illustrate with ProntoQA-95 (gold = B): a chain of universal
implications resolves to ``Stella is not feisty.''

\paragraph{Qwen-4B.} The first tool call is itself malformed at the
envelope level: the model wraps its payload as
\verb!{"arguments":"{\"smt_code\":\"...\"}"}!, which the harness's
JSON-strict tool parser cannot decode, so the next \texttt{tool} message
is empty. Inside the inner payload the SMT-LIB is also incorrect:

\begin{lstlisting}
(set-logic QF_LIA)
(declare-const IsFeisty Pred)
(declare-const IsVumpus Pred)
(declare-const Stella Bool)
(assert (= (IsVumpus Stella) True))
(assert (forall ((x Pred))
  (=> (= (IsVumpus x) True)
      (= (IsDumpus x) True))))
\end{lstlisting}

\noindent Several things are wrong: \texttt{Pred} is not a sort,
\texttt{IsVumpus} is declared as a constant but applied as a function,
\texttt{Stella} is a \texttt{Bool} but used as an individual, and
\texttt{True} appears as a value rather than the SMT-LIB literal
\texttt{true}. Over the next eight turns the model loses track of the
original problem and cycles through
\verb!(set-logic QF_LIA)!,
\verb!(set-logic QF_BV)!,
\verb!(set-logic QF_NIA)!, and
\verb!(set-logic QF_ARITH)!, the last of which is not a recognized
logic name. It exhausts its 10-turn budget without emitting
\texttt{finish}.

\paragraph{Sequent-Prover-4B.} The first tool call is a first-order encoding with a
sort for animals and one predicate per type:

\begin{lstlisting}
(declare-sort Animal)
(declare-fun vumpus  (Animal) Bool)
(declare-fun dumpus  (Animal) Bool)
(declare-fun rompus  (Animal) Bool)
;; ... (one declare-fun per category) ...
(declare-const Stella Animal)

(assert (forall ((x Animal))
  (=> (vumpus x) (dumpus x))))
;; ... the full implication chain ...
(assert (forall ((x Animal))
  (=> (zumpus x) (not (feisty x)))))
(assert (vumpus Stella))
(assert (feisty Stella))   ; proof by contradiction
(check-sat)                ; --> unsat
\end{lstlisting}

\noindent The first pass surfaces two ordering errors (\texttt{yumpus}
and \texttt{floral} are used before they are declared). The Sequent-Prover-4B model
reads the Z3 error, hoists the missing \texttt{declare-fun}s to the top,
re-runs, and finishes in four turns. The base model, given the same
kind of error, could not localize the fix and drifted into unrelated
probes.

\subsection{Capability 2: First-order over propositional encoding}
\label{sec:fol}

When a problem quantifies over individuals, a propositional encoding
(one boolean per ground proposition) often forces the model to fill in
unstated relations by hand. The base model does this routinely; Sequent-Prover
does not.

\paragraph{Example: FOLIO-1361 (gold = Uncertain).} Premises describe
animals near a campus and ask whether ``Rockie is a turtle.'' The
critical step is to notice that the premises constrain Rockie to be a
rabbit \emph{or} a squirrel but say nothing about whether rabbit and
turtle are disjoint categories.

\paragraph{Qwen-4B.} Declares one \texttt{Bool} per (Rockie, property)
pair, with no animal sort. Lacking a way to express ``the only animals
spotted are rabbits and squirrels'' as a universal, it instead injects
two disjointness axioms that are nowhere in the premises:

\begin{lstlisting}
; Assumption: Rabbits and turtles are mutually exclusive
(assert (=> RockieRabbit (not RockieTurtle)))

; Assumption: Squirrels and turtles are mutually exclusive
(assert (=> RockieSquirrel (not RockieTurtle)))
\end{lstlisting}

\noindent With those injected axioms, the negation of the conclusion
becomes \texttt{unsat} and the base model answers ``False'', but the
false answer is caused by the hallucinated premise, not by the actual
problem.

\paragraph{Sequent-Prover-4B.} Uses a sort \texttt{Animal} and universally quantified
predicates over animals; the conditions in the prompt translate
mechanically. We discuss Sequent-Prover's full reasoning trajectory under
Capability~6 below, because it turns on the same diagnostic move used in
another problem.

\subsection{Capability 3: Bidirectional satisfiability\\for ternary verdicts}
\label{sec:bidirectional}

When the verdict space is ternary (True / False / Uncertain) or
four-valued (entailment / contradiction / neutral / self-contradiction),
a single satisfiability query is not enough. The textbook idiom is two
queries: assert the conclusion, then assert its negation. The base model
asks one and stops; Sequent-Prover-4B asks both.

\paragraph{Example: FOLIO-476 (gold = False).} ``Tom will rent at the
Olive Garden,'' under premises constraining the deposit to be at least
the monthly rent (\$2000) and renting to require deposit
$\leq$\$1500.

\paragraph{Qwen-4B.} Spends three turns repairing surface syntax
(\texttt{Boolean}$\rightarrow$\texttt{Bool}; \texttt{->} to \texttt{implies};
\texttt{(= oliveWillRent False)}$\rightarrow$\texttt{(not oliveWillRent)}).
Its single satisfiability query is on the negation of the conclusion,
which is \texttt{sat}, so it concludes ``the conclusion does not
necessarily follow'' and answers \textbf{Uncertain}. It never asks the
dual question.

\paragraph{Sequent-Prover-4B.} Runs both queries:

\begin{lstlisting}
;; First: can the conclusion FAIL under the premises?
(assert (not (and allowed_to_move_in_with_fluffy
                  (<= deposit_at_olive 1500.0))))
(check-sat)             ;  sat   ==> not forced True

;; Second: can the conclusion HOLD under the premises?
(assert (and allowed_to_move_in_with_fluffy
             (<= deposit_at_olive 1500.0)))
(check-sat)             ; unsat  ==> forced False
\end{lstlisting}

\noindent (sat, unsat) is the signature of a forced-false conclusion.
Sequent-Prover-4B finishes with \textbf{False} and the explanation reproduces the
informal argument: ``\$2000 $>$ \$1500, so the deposit condition cannot
be satisfied at the Olive Garden.'' Both models know
\texttt{check-sat}. Sequent-Prover-4B differs in its choice of
\emph{which experiment to run next}.

\subsection{Capability 4: Declarative encoding via datatypes}
\label{sec:datatypes}

For problems with explicit finite enumerations (LSAT analytical
reasoning, ZebraLogic), the canonical idiom is
\texttt{declare-datatypes} for each finite domain and a function from
one to another. 

\paragraph{Example: AR-LSAT-82 (gold = E).} Seven sales reps assigned to
three zones, with same-zone pairings and an XOR-style
either-but-not-both rule. ``If more reps work in Zone~1 than in
Zone~3, which one of the following \emph{could} be true?''

\paragraph{Qwen-4B.} Encodes zones as unbounded integers and emits a wall
of errors over all ten turns. The actual bugs we observe across calls
are: bare \texttt{<= 1 zone\_Kim 3} expressions at top level (outside
any \texttt{assert}); ad-hoc \texttt{[zone\_Kim]} bracket syntax;
\texttt{(select zone\_Kim 1)} on a \texttt{Bool} as if to extract an
indicator; the conjunction
\texttt{(and (= zone\_Kim 1) (= zone\_Kim 2) (= zone\_Kim 3))} used
where the intent is disjunction; \texttt{(if ...)} instead of
\texttt{(ite ...)}; and a multi-variable declare,
\texttt{(declare-const zK zM zP zQ zS zT zU Int)}, which is not legal
SMT-LIB. None of these get repaired; each turn re-issues a near-copy of
the previous one. The model never reaches the option-testing stage.

\paragraph{Sequent-Prover-4B.} First turn uses two algebraic datatypes and an abstract
assignment function:

\begin{lstlisting}
(declare-datatypes ()
  ((Rep Kim Mahr Parra Quinn Stuckey Tiao Udall)))
(declare-datatypes ()
  ((Zone Zone1 Zone2 Zone3)))
(declare-fun zone (Rep) Zone)

(assert (xor (= (zone Parra) Zone1)
             (= (zone Tiao)  Zone1)))     ; rule 1
(assert (xor (= (zone Tiao)  Zone2)
             (= (zone Udall) Zone2)))     ; rule 2
(assert (= (zone Parra) (zone Quinn)))    ; rule 3
(assert (= (zone Stuckey) (zone Udall)))  ; rule 4

(define-fun count1 () Int
  (+ (ite (= (zone Kim)  Zone1) 1 0)
     (ite (= (zone Mahr) Zone1) 1 0)
     ;; ... ite per rep ...
     ))
\end{lstlisting}

\noindent The first pass writes \texttt{(not = =)} (the constraint is
``exactly one of X, Y'') and gets a Z3 arity error; the next turn
replaces the offending subtree with \texttt{xor}. This is the same
local-edit recovery pattern as in Capability~1.

\subsection{Capability 5: Push/pop for multi-option queries}
\label{sec:pushpop}

LSAT analytical-reasoning items ask ``which of the following could /
must be true?'' This is a sequence of independent satisfiability tests
on a shared base theory, the SMT idiom is \texttt{push}/\texttt{pop}.

\paragraph{Example: AR-LSAT-60 (gold = E).} Four riders and four bicycles
across two days, with same-day distinctness and cross-day pairing.
``If Theresa tests G on day~2, which one of the following \emph{must}
be true?''

\paragraph{Qwen-4B.} Across all ten turns the same family of errors
recurs. Seven of the ten calls contain
\texttt{(assert (= T1 H) or (= T2 H))}, with \texttt{or} as a
top-level token outside the \texttt{assert}, which Z3 reports as
``invalid assert command, \texttt{)} expected.'' The base model treats
the error as a request to reformat the body (one call collapses the
entire program onto a single line; another swaps in unbounded integer
ranges), but never rewrites the offending node as the prefix expression
\texttt{(or (= T1 H) (= T2 H))}. It never reaches the option-testing
stage, and \texttt{extraction\_method = none}.

\paragraph{Sequent-Prover-4B.} A single first-turn body emits five
\texttt{push}/\texttt{pop} blocks, one per answer choice. Each block
asserts the \emph{negation} of the choice: the canonical pattern for
``must be true'' (if the negation is unsat, the choice must hold).
\texttt{(echo ...)} labels the output so the five results are
self-describing.

\begin{lstlisting}
(declare-datatypes () ((Rider R Seamus T Yuki)))
(declare-datatypes () ((Bike  F G H J)))
(declare-fun d1_rider (Bike) Rider)
(declare-fun d2_rider (Bike) Rider)

(assert (distinct (d1_rider F) (d1_rider G)
                  (d1_rider H) (d1_rider J)))
;; ... base theory ...
;; Yuki's day-1 bike is Seamus' day-2 bike:
(assert (exists ((b Bike))
  (and (= (d1_rider b) Yuki)
       (= (d2_rider b) Seamus))))

(echo "Testing (A): Reynaldo tests H, day 1")
(push) (assert (not (= (d1_rider H) R)))
       (check-sat) (pop)
;; ... (B), (C), (D) similarly ...
(echo "Testing (E): Yuki tests H, day 2")
(push) (assert (not (= (d2_rider H) Yuki)))
       (check-sat) (pop)
\end{lstlisting}

The single Z3 reply ends with ``Testing (E) \dots\ unsat,'' the unique
negation that is unsat, so (E) is the choice that must hold. Sequent-Prover
finishes in three iterations, having issued one comprehensive query
instead of five separate ones.

Three sub-skills are bundled into this first turn: (i)~scaffolding with
\texttt{(echo ...)} so the Z3 output is self-describing; (ii)~knowing
\emph{which} direction to negate for ``must'' vs.\ ``could'' questions;
(iii)~knowing that \texttt{push}/\texttt{pop} make the five queries
independent. Note also the encoding choice: inverting the natural
direction to \texttt{d1\_rider : Bike $\rightarrow$ Rider} lets the
day-1 distinctness collapse to a single \texttt{distinct} call
instead of six pairwise inequalities, and lets the cross-day pairing be
expressed as a one-variable existential.

\subsection{Capability 6: Diagnosing inconsistent\\premise sets}
\label{sec:selfcontra}

A further capability is noticing that
\emph{both} directions of a hypothesis are unsatisfiable, and inferring
from that signature that the premise set itself is inconsistent. This
move appears in two superficially unrelated transcripts.

\paragraph{Example A: Self-Contradiction} We used an example where there were twelve
atomic facts and twelve quantified rules over seven named people. The
base model spends ten turns rewriting an 80-line SMT body, hitting the
same family of errors each time (missing \texttt{Harold\_brainy}
declaration, mis-parenthesized rules~4 and~9). At turn~22 the harness
sends a follow-up nudge, ``Z3 returned a clean UNSAT result, so now
emit the finish tool call'', but the base ignores it and issues yet
another \texttt{z3\_solve} on turn~23.

Sequent-Prover-4B arrives at the verdict in three steps. First, it tests both
directions of the hypothesis:

\begin{lstlisting}
;; (a) Is the hypothesis decidable?
(push) (assert R_huge)
       (check-sat) (pop)      ;  unsat
(push) (assert (not R_huge))
       (check-sat) (pop)      ;  unsat
\end{lstlisting}

\noindent Both \texttt{unsat} is the diagnostic signature of an
inconsistent premise set. Sequent-Prover-4B confirms by querying the premises alone:

\begin{lstlisting}
;; (b) Are the premises consistent?
;;   ... (only premises asserted) ...
(check-sat)                   ;  unsat
\end{lstlisting}

\noindent and finishes with \texttt{self\_contradiction}. The
explanation it produces names the diagnostic explicitly: ``Z3 returns
UNSAT for the premises alone, and both \texttt{R\_huge} and
\texttt{(not R\_huge)} are UNSAT when added to the premises. This means
the premises contain an inherent contradiction.''

\paragraph{Example B: FOLIO-1361, revisited.} Recall that the base
model on this problem (\S\ref{sec:fol}) injected ad-hoc disjointness
between rabbits, squirrels, and turtles and got an unsound ``False.''
Sequent-Prover's trajectory uses the same diagnostic move as Example~A:

\begin{enumerate}[leftmargin=*,itemsep=0pt,topsep=2pt]
\item No injected axioms; both directions of \texttt{(turtle Rockie)}
return \texttt{sat}. (Sequent-Prover-4B does not yet realize this \emph{is} the
``Uncertain'' answer.)
\item Adds disjointness axioms; the negation goes \texttt{unsat},
suggesting the conclusion is forced True.
\item Runs both push/pop queries on the disjointness-augmented theory:
\emph{both} are \texttt{unsat}.
\item Recognizes both-\texttt{unsat} as a contradiction signature,
checks the augmented premises alone, gets \texttt{unsat}.
\item Concludes the disjointness axioms are not licensed; removes them.
\item Re-runs both directions: \texttt{sat} and \texttt{sat}.
\item Finishes with \texttt{Uncertain} and a verbal argument: ``The
premises do not state that rabbits and turtles are disjoint
categories.''
\end{enumerate}

\noindent The same meta-strategy, both UNSAT $\Rightarrow$ suspect
your own axioms, recovers the correct answer on a problem where the
base model is led astray by the axioms it injected. This strategy
spans several solver calls, and the model learns it as a policy over
them.

\subsection{Capability 7: Error text as actionable feedback}
\label{sec:recovery}

The recovery patterns we observe in Sequent-Prover-4B follow a recurring structure:
read the Z3 error string, locate the named node in the previous body,
and patch it without disturbing the rest of the program. The base model
either re-emits a near-identical body or escalates the change in
unhelpful directions (switching logics, collapsing onto one line).
Three mini-patterns:

\begin{description}[leftmargin=*,style=nextline,itemsep=2pt,topsep=2pt]
\item[ProntoQA-95.] Error: ``unknown constant \texttt{yumpus}
(Animal).'' Fix: move \texttt{(declare-fun yumpus (Animal) Bool)} above
its first use. Nothing else changes.
\item[FOLIO-476.] Error: ``unknown sort \texttt{Boolean}.'' Sequent-Prover-4B patches
each occurrence to \texttt{Bool} in one turn. The base model takes two
extra turns to also re-discover that \texttt{->} is not a keyword
(use \texttt{=>} or \texttt{implies}) and that the constant is
lower-case \texttt{false}.
\item[AR-LSAT-82.] Error: ``Wrong number of arguments (2) passed to
function \texttt{not}.'' Fix: replace the entire
\texttt{(not (= ...) (= ...))} subtree with
\texttt{(xor (= ...) (= ...))}, which is both shorter and correct.
\end{description}

\noindent The base model knows the SMT-LIB tokens. It does not use the
parse tree, symbol table, or scopes to locate an error, and it retries
the whole program instead.

\label{sec:appendix}

\end{document}